# A Survey on Content-aware Video Analysis for Sports

Huang-Chia Shih, *Member, IEEE*

*Abstract*—Sports data analysis is becoming increasingly large-scale, diversified, and shared, but difficulty persists in rapidly accessing the most crucial information. Previous surveys have focused on the methodologies of sports video analysis from the spatiotemporal viewpoint instead of a content-based viewpoint, and few of these studies have considered semantics. This study develops a deeper interpretation of content-aware sports video analysis by examining the insight offered by research into the structure of content under different scenarios. On the basis of this insight, we provide an overview of the themes particularly relevant to the research on content-aware systems for broadcast sports. Specifically, we focus on the video content analysis techniques applied in sportscasts over the past decade from the perspectives of fundamentals and general review, a content hierarchical model, and trends and challenges. Content-aware analysis methods are discussed with respect to object-, event-, and context-oriented groups. In each group, the gap between sensation and content excitement must be bridged using proper strategies. In this regard, a content-aware approach is required to determine user demands. Finally, the paper summarizes the future trends and challenges for sports video analysis. We believe that our findings can advance the field of research on content-aware video analysis for broadcast sports.

## I. INTRODUCTION

RESEARCH interest in sports content analysis has increased substantially in recent decades, because of the rapid growth of video transmission over the Internet and the demand for digital broadcasting applications. The massive commercial appeal of sports programs has become a dominant focus in the field of entertainment. Research on big data analytics has attracted much attention to machine learning and artificial intelligence techniques. Accordingly, content analysis of sports media data has garnered attention from various studies in the last decade. Sports data analysis is becoming large scale, diversified, and shared. The most pressing problem currently is how to access the most important information in a short time.

Because of the massive demand for sports video broadcasting, many enterprises such as Bloomberg, SAP, and Vizart employ sports content analytics. Content analysis with big data has used become a major emerging industry. In offline service, historical records can be used to analyze video content through machine learning. In online service, discovered latent knowledge can be for real-time tactic recommendation. In recent years, many books have contributed to the content analysis of statistics for baseball [1]–[3] and basketball [4]–[6].

At present, several sports intelligence systems and content analytics have been developed and applied:

1) Panasonic incorporated SAP SE to develop a video-based sports analytics and tracking system. Match Insights is an analytics prototype SAP developed with the German Football Association for the soccer World Cup 2014 in Brazil [7].
2) Vizrt, short for VisualiZation (in) real time, is a Norwegian company that provides content production, management, and distribution tools for the digital media industry. Its products include applications for creating real-time 3D graphics and maps, visualizing sports analyses, managing media assets, and obtaining single workflow solutions for the digital broadcast industry. Vizrt has a customer base in more than 100 countries and approximately 600 employees distributed at 40 offices worldwide [8], [9].
3) PITCHf/x data set is a public resource presented by MLBAM [10] and Sportvision [11]. Brooks Baseball [12] makes systematic changes to this data set to improve its quality and usability. They manually review the Pitch Info by using several parameters of each pitch's trajectory and validate the parameters against several other sources such as video evidence (e.g., pitcher grip and catcher signs) and direct communication with on-field personnel (e.g., pitching coaches, catchers, and the pitchers themselves). The default values of the trajectory data are slightly altered to align them more closely with the real values.
4) Sportradar [13], a Swiss company, focuses on collecting and analyzing data related to sports results by collaborating with bookmakers, national soccer associations, and international soccer associations. Their operating activities include the collection, processing, monitoring, and commercialization of sports data, which result in a diverse portfolio of sports-related live data and digital content.

### A. Surveys and Taxonomies

Since 2000, sports video analysis has continually drawn research attention, leading to a rapid increase in published work and surveys in this field. In [14], a preliminary survey of sports video analysis was conducted to examine diverse research topics such as tactic summarization, highlight extraction, computer-assisted referral, and content insertion. Kokaram *et al.* [15] demonstrated the development trends in the topics of sports-related indexing and retrieval. They identified two broad classes of sports, court and table sports and field sports. However, this was not a favorable classification for content analysis. An alternative classification criterion, the time- or point-driven structure in the progress of a game, was then considered. Sports such as baseball and soccer are time driven because the high-excitement events occur sparsely and randomly. Conversely, point-driven sports consist of regular events and constructs within a domain-specific scenario. In this type of sports, such as tennis and snooker, highlight events yield particular points in the game structure. Different from the viewpoint of Kokaram *et al.* [15], Santiago *et al.* [16] focused on the application of team tracking techniques to sports games. They categorized the reviewed systems into —two classes, intrusive and nonintrusive. In intrusive systems, tags or sensors are placed on the targets, whereas nonintrusive systems introduce no extra objects to the game environment, instead using vision as the main sensory source and employing image processing techniques to perform player and team tracking. Intrusive systems are typically not considered highly mature in providing high-level information. A domain-specific survey of soccer video analysis was presented in [17]. The authors reviewed studies on soccer video analysis at three semantic levels of interpretation: video summarization, provision of augmented information, and high-level analysis. In addition, the computational problems at these three semantic levels were discussed. First, the two main tasks of video summarization include extracting the most interesting parts of the video and omitting the less interesting parts. Studies on event detection and highlight extraction were reviewed from the viewpoints of features, models, and classifiers. Second, studies on provisions of augmented information for presenting the game status to viewers were discussed. Third, extension of the level of video summarization was proposed. Considering the interactions between objects, more complex analyses, such as team statistical analysis and real-time event analysis, can be conducted. Specifically, an overview of automatic event detection in soccer games was presented [18]; this overview involved simply classifying approaches into a hierarchical structure on the basis of their analysis levels (i.e., low, middle, and high). Similarly, Rehman and Saba reviewed feature extraction for soccer video semantic analysis [19]. In addition, audio and video feature extraction methods and their combination with textual methods have been investigated. The data sources, methodologies, detected events, and summarization applications used in event-oriented soccer video summarization have been compared. **Table I** shows a comparison of previous surveys.

TABLE I
KEY ISSUES OF PREVIOUS SURVEYS

| Works | Years | Sport Genres | Major subjects | Minor subjects | Goal | # of ref. |
|---|---|---|---|---|---|---|
| [14] | 2003 | Comprehensively | Event | Tactics analysis, highlight extraction, tracking, computer-assisted refereeing, content insertion | Sports Video Analysis | 28 |
| [15] | 2006 | Snooker, tennis, badminton, cricket, soccer, American football, baseball, sumo | Object, Event, Context | Feature extraction, event detection, video summarization, indexing, and retrieval | Sports content retrieval Court/table sports vs. field sports Bridging the semantic gap | 54 |
| [16] | 2010 | Football, handball, basketball, squash, skating, indoor soccer | Objects | Team tracking | Intrusive vs. nonintrusive systems Indoor vs. outdoor sports | 43 |
| [17] | 2010 | Soccer | Event | Video summarization, provision of augmented information, high-level analysis | Low-level visual features vs. high-level semantic analysis | 89 |
| [18] | 2011 | Soccer | Object, Event | Highlights, summarization, event detection | General models for soccer analysis. | 28 |
| [19] | 2014 | Soccer | Event | Feature extraction, highlight event detection, video summarization | Audio and video features, textual method integration | 56 |

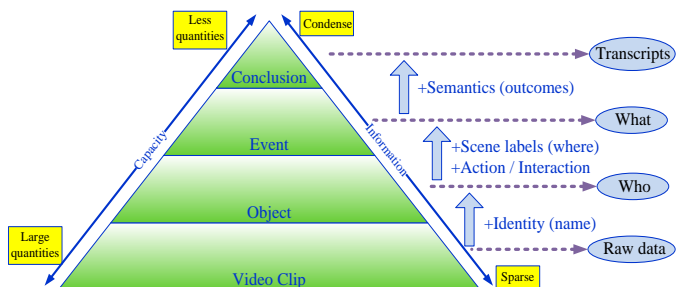

**Fig. 1**. Content Pyramid: a hierarchical content model.

In summary, previous surveys have focused on the methodologies of sports video analysis from the spatiotemporal viewpoint instead of a content-based viewpoint. Few of these studies considered semantics. The current study adopted a different framework, developing a proper and deeper interpretation of content-aware sports video analysis. Specifically, we present the insight offered by research into the structure of content under different scenarios. On the basis of this insight, we provide an overview of the themes particularly relevant to the research on content-aware systems for sports.

*B. Scope and Organization of the Paper*

A video analysis system can be viewed as a content reasoning system. In this paper, we review the developments in sports video analysis, focusing on content-aware techniques that involve understanding and arranging the video content on the basis of intrinsic and semantic concepts. We focus on the video content analysis applied in sportscasts over the past decade from three aspects—fundamentals and general review, a content hierarchical model, and challenges and future directions.

- First, we introduce the fundamentals of content analysis, namely the concept of the content pyramid, sports genre classification, and the overall status of sports video analytics.
- Second, we review state-of-the-art studies conducted in this decade on the content hierarchical model (i.e., content pyramid). The information used to represent high-level semantic knowledge can be divided into three groups: object-, event-, and context-oriented groups.
- Third, we review the prominent challenges identified in the literature and suggest the promising directions for future research on sports video content analysis.

The remainder of this paper is organized as follows. In Section II, we discuss the fundamentals of content-aware video analysis for sports. A systematic and detailed review of state-of-the-art studies according to three content awareness levels is presented in Section III. In Section IV, we present the challenges and promising future directions regarding to the content-aware sports video analysis for expert users. A conclusion is drawn in the final section.

## II. FUNDAMENTALS OF CONTENT-AWARE SPORTS VIDEO ANALYSIS

*A. Content Pyramid*

In developing a content-aware retrieval system, the retrieval process must be designed according to users' intention. Several

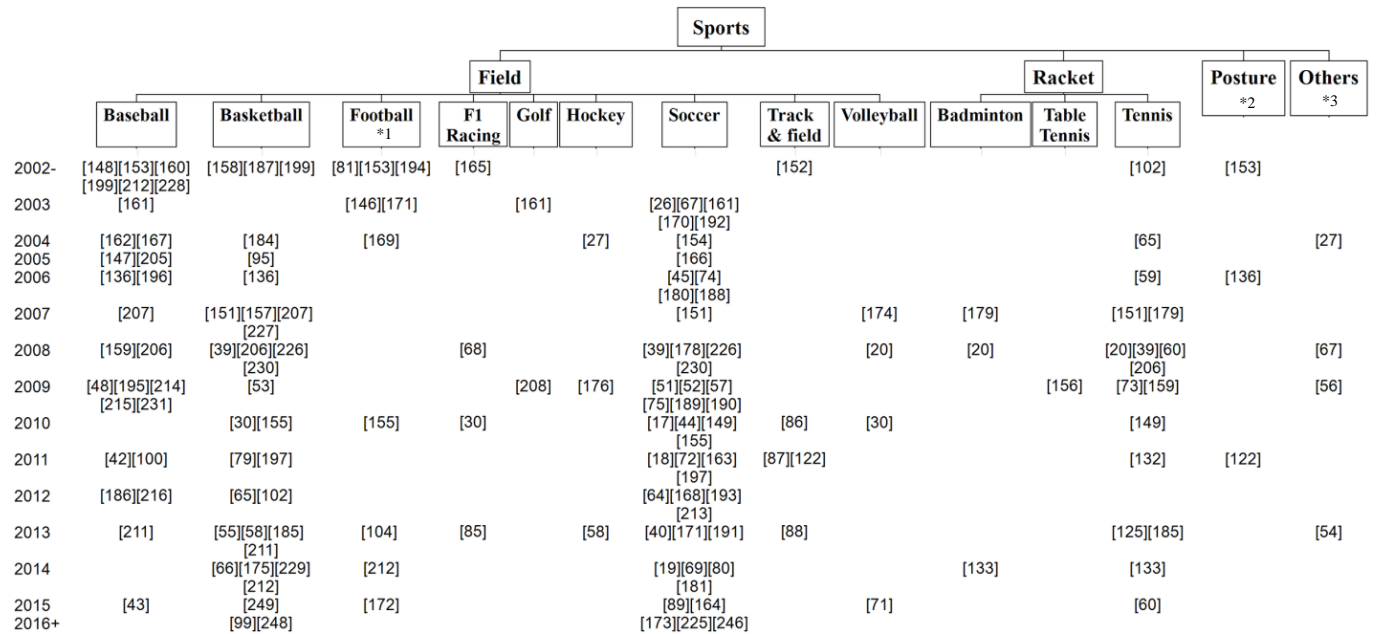

*1: American football: [81], [153], [169], [171], [194]; Gaelic football: [146]; Australian football: [155]; No specified: [104], [212], [238]
*2: Sumo Wrestling: [153]; Figure Skating: [136]; Diving: [122]
*3: Swimming: [27], [54]; Skiing: [68]; Speed Skating: [56]

**Fig. 2**. Types of sports classification.

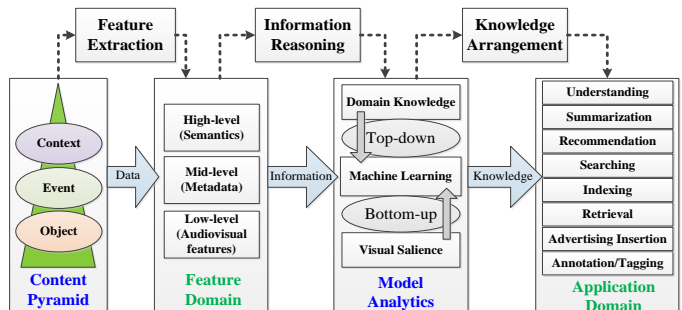

**Fig. 3**. Paradigm of content-aware video analysis.

sports video analysis systems have focused on managing visual features in the spatial domain. For example, Han *et al.* [20] introduced a general framework for analyzing broadcast court-net sports videos from the viewpoints of the pixel, object, and scene levels. Using this framework, a group of predefined events at different levels can be identified in advance. According to the semantic significance, the video content was divided into four layers (**Fig. 1**): video, object, action, and conclusion layers [21], [22]. The volume of the layer denotes the quantity of the implied concepts. The compactness of information deceases from the top down. The concept of the content pyramid is used to analyze context concerns for a video entity. Each layer of the content hierarchy represents various key components for characterizing the video content. The video layer consists of video clip frames, each of which consists of a video clip tag and raw video data. The object class layer consists of object frames, which represent the key objects in the video. In each object frame, an object tag and pointers link each key object to the corresponding video clips. An object that has a particular posture or movement or interacts with other objects yields an action or interaction tag. The event class layer consists of event frames, which represent the action of the key object. Actions combined with scene information constructs an event tag. Each event frame consists of an event tag and an object tag, representing the related action or interaction among multiple objects. The top layer is the conclusion layer, which consists of conclusion frames representing the semantic summarization of the video sequence. Each conclusion frame consists of the event tags and corresponding results. A game summary is drafted according to transcripts and outcomes from events.

*B.  Sports Genre Categorization*

Recently, the proliferation and tremendous commercial potential of sports videos have strongly indicated the need for relevant applications. The technique of genre classification has become a common preliminary task for managing sports media. **Fig. 2** shows a tree structure of sports genre classifications, presenting the reviewed papers by year of publication. The sports typically broadcasted on TV can be classified into three categories: field, posture, and racket. As **Fig. 2** shows, more than 80% of papers addressed baseball, basketball, soccer, and tennis. This figure shows only papers that focused on fewer than 4 types of sports; other papers are included in a general category. Typically, golf is considered both a field sport and a posture sport. Some track-and-field sports such as the long jump and pole vaulting are posture sports, whereas other sports such as the hammer throw and javelin throw are classified as racket sports. The remainder of this section focuses on reviewing methods for sports genre classification.

Brezeale *et al.* [23] reviewed studies on video classification and identified the use of features in three modalities: text, audio, and visual. Videos can be classified according to genres, such as movies, news, commercials, sports, and music, and subgenres. In the last decade, various sports genre classification methods based on different classifiers such as HMMs [24]–[26], naïve Bayesian classifiers (NBCs) [26], decision trees [27], and support vector machines (SVMs) [28], [29], have been presented. Schemes combining different classifiers have also been considered. You *et al.* [30] combined NBC and HMM algorithms to classify ball match video genres and detect events in football, basketball, and volleyball. Duan *et al.* [31] combined an NBC and SVM to categorize game videos into five typical ball sports, namely tennis, basketball, volleyball, soccer, and table tennis. Zhang *et al.* [32] adopted the bag-of-visual-words (BoW) model with k-nearest neighbor classifiers to identify video genres. In their method, to achieve a generic framework for analyzing sports videos, an unsupervised probabilistic latent semantic analysis-based approach was used to characterize each frame of a video sequence into one of the following four groups: close-up view, mid-view, long view, and outer-field view.

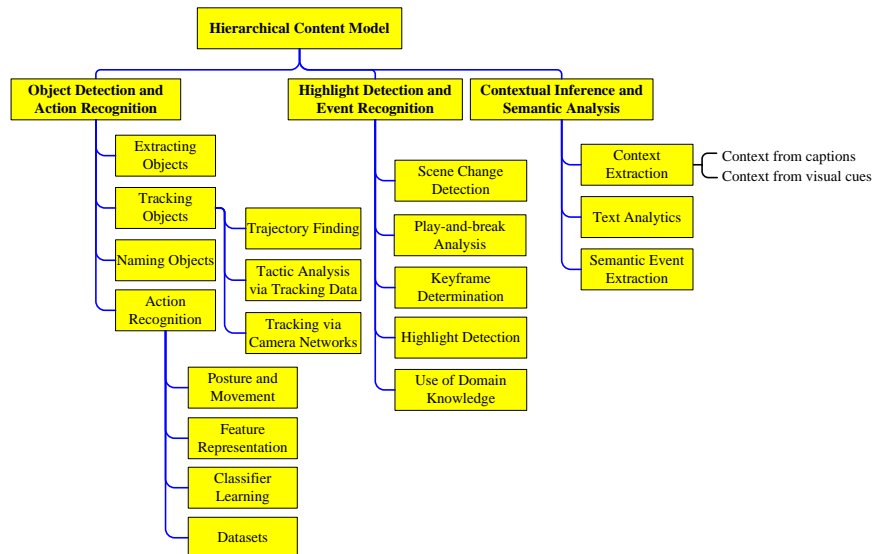

**Fig. 4**. Taxonomy of surveys with different semantic level applications

For mobile video application, Cricri *et al.* [33] used a multimodal approach with four auxiliary data sets—sensor-embedded mobile devices, spatial and spatiotemporal visual information, and audio data, for confirming the sports type. To integrate all possible modality combinations in fusion models, the auxiliary input data can be used for video type classification. For home video application, Sugano *et al.* [34] extracted the salient low-level features from unedited video. The random forest [35] and the bagging [36] algorithms were used to classify the home videos into five genres, namely sports, travel, family and pets, event, and entertainment.

### C. Overview of Sports Video Analysis

Sports video analysis aims to extract critical information from video content. **Fig. 3** provides an overview of sports video analysis. The general procedure of a content-aware video analysis system includes feature extraction, information reasoning, and knowledge arrangement. Based on the concept of the content pyramid, content-aware video analysis can be performed using different level approaches. Information reasoning is used to understand low-, mid-, and high-level video data. Using model analytics, the information can be converted to knowledge. In other words, model analytics entails transforming data in the feature domain to knowledge in the application domain. The methodologies of model analytics can be categorized into two classes: top-down and bottom-up.

The top-down methodology assumes that the structure of the model is given, whereas the bottom-up methodology starts with visual salience information treated as source features through machine learning to obtain high-level knowledge.

### III. SURVEYS OF CONTENT-AWARE VIDEO ANALYSIS WITH SEMANTIC LEVELS

According to the structure of the content pyramid, we surveyed state-of-the-art techniques from the aspect of semantic level, namely highlight detection and event recognition, object detection and action recognition, and contextual inference and semantic analysis. The surveyed techniques are categorized in **Fig. 4**.

### A. Object Detection and Action Recognition

Object-oriented sports content analysis is one of the most active research domains. This approach has been successfully applied for action recognition using intraobject posture analysis and interobject event determination. For single-object application, an accurate body posture and accurate movement are required to obtain the desired performance in sports such as diving, tumbling, skating, ballet, golf, and track-and-field games. For two-object application, the desired performance in sports such as boxing, sumo, wrestling, tennis, table tennis, and badminton is obtained using between-object interactions and that in sports such as soccer, rugby, hockey, basketball, baseball, and volleyball is obtained using between-group interactions. For multiple-object application, many studies focus on the group-to-group interaction for recognizing team tactics.

*1) Extracting Objects*

In sports videos, the foreground object plays a crucial role in the event scenario. Its actions and interaction with other objects can form a particular event. The objects in a video frame, such as auxiliary blobs, figures, and texts, can be either moving or stationary and either natural or superimposed. To convert low-level media features to high-level semantic labels,

Naphade *et al.* [37] created a terminology named probabilistic multimedia objects (Multijects). To verify the dependency between intraframe and interframe semantics, Multijects are combined with a graphical network that captures the co-occurrence probabilities of Multijects, named Multinet. A robust automatic video object (VO) extraction technique was presented in [38]. The authors of that study reported that the dynamic Bayesian network (DBN) framework can facilitate attaining a more detailed description of VO characteristics compared with an HMM.

In a soccer game, the player and ball positions constitute the most critical information. Pallavi *et al.* [39] proposed a hybrid technique for detecting the ball position in soccer videos. This technique first classifies a shot as a medium or long shot, and motion- and trajectory-based approaches are then used for detecting medium and long shots, respectively. This method does not require any preprocessing of the videos and is rather efficient. A framework based on the probabilistic analysis of salient regions was introduced in [40]. Player detection and tracking in broadcast tennis videos were presented in [41]. In [42], for a baseball game, an object segmentation algorithm was first used to retrieve the pitcher's body, and a star skeleton of the pitcher was then used to recognize the pitching style. The authors presented automatic generation of game information and video annotation through a pitching style recognition approach. Similarly, an algorithm for pitch-by-pitch video extraction proposed in [43] incorporated a pitcher localization technique, a pitch speed recognition scheme, and a motion degree measurement scheme for accurately identifying pitches.

However, there is a difficulty in extracting objects perfectly. A scene normally includes multiple objects in the playfield. The objects can partially or completely occlude each other or self-occlude because of the capturing angle. Hamid *et al.* [44] used multiple cameras to detect objects. In addition, 3D information has been used to estimate player and ball positions in broadcast soccer videos [45], [46]. This enabled the authors to measure the aerial positions without referencing the heights of other objects. Moreover, using the Viterbi decoding algorithm, they could determine the most likely ball path by considering consecutive frames. Recently, Li *et al.* [47] introduced the object bank for representing an image by using a high-level image representation to encode object appearance and spatial location information.

Furthermore, studies have developed processes for quantifying the importance of VOs by using the object attention model [48] and visual attention model [49], [50]. On the basis of such processes, the importance score of an object can be used to infer the importance of a frame and even a video clip. If a frame contains more high-attention objects with a high-interest contextual description, it is highly probable that the frame has higher excitement score. Regarding an event-based analysis scenario, the score of each frame is derived on the basis of the relevant key frames that are located in an identical event.

*2) Tracking Objects*

Object tracking is a crucial technique in sports content analysis. It aims to localize the objects in a video sequence and provides diverse applications in video surveillance, HCI, and visual indexing. In sports, the object tracking technique is used to observe the continuing activities of objects (e.g., players and ball) in the playfield. However, the camera motion is a major limitation in detecting and tracking objects. Khatoonabadi *et al.* [51] applied a region-based detection algorithm for eliminating fast camera motion effects in goal scenes and tracking soccer players. When the region-based algorithm is used to track players, either the template matching method or split-and-merge approach is applied for occlusion reasoning between players. However, players' positions can be misdetected in the split-and-merge approach. Liu *et al.* [52] presented a method for detecting multiple players, unsupervised labeling, and efficient tracking in broadcast soccer videos. The framework involves a two-pass video scan, which first learns the color of the video frame and then tests the appearance of the players. The method is generally effective, except for cases involving video blur and sudden motion. Kristan *et al.* [53] presented an algorithm for tracking multiple players in an indoor sporting environment. Considering a semicontrolled environment with certain closed-world assumptions, the algorithm relies on a particle filter. The multiplayer tracking process involves background elimination, local smoothing, and management of multiple targets by jointly inferring the players' closed worlds. In addition, Sha *et al.* [54] presented an automatic approach for swimmer localization and tracking. The approach enables large-scale analysis of a swimmer across many videos. A swimming race is divided into six states: a) start, b) dive, c) underwater, d) swim, e) turn, and f) end. A multimodal approach is employed to modify the individual detectors and tune to each race state. However, the location information cannot present more detailed context of the game. To solve this problem, long-term observation is required for tracking object movement.

*a) Trajectory finding*

A trajectory is the path along which a moving object passes through space as a function of time. Robust solutions to trajectory-based techniques have applications in domains such as event reasoning and tactic analysis. Chakraborty *et al.* [55] presented a real-time trajectory-based ball detection and tracking framework for basketball videos. Liu *et al.* [56] proposed a computer vision system that tracks high-speed nonrigid skaters over a large area. The input video sequence is fed into a registration subsystem and a tracking subsystem, and the results from both subsystems are used to determine the spatiotemporal trajectory, which can be used for event detection and sports expert analysis. However, tracking performance is lacking when skaters move in groups during long and continual complete occlusions. A possible solution was presented by Miura *et al.* [57]. They attempted to estimate ball routes and movements by tracking overlaps with players and lines in order to recognize certain scenes. When an overlap occurs, the spatiotemporal relationship between the ball and the object is examined. Ball existence near each route candidate is then examined to eliminate potential routes and determine the most likely route. Similarly, Liu *et al.* [58] presented context-conditioned motion models incorporating complex interobject dependencies to track multiple players in team sports over long periods. Yan *et al.* [59], [60] have presented a robust tennis ball tracking method based on a multilayered data association with graph-theoretic formulation. More recently, Zhou et al. [61] proposed a two-layered data association approach for tennis ball tracking that was identical to Yan's method, but did not consider Yan's tracklet linkage, which may cause tracking failure. For addressing short-term misdetection, Wang et al. [62], [63] introduced a more efficient dense trajectory. They combined trajectory shape, appearance, and motion information to compute motion boundary descriptors along dense trajectories.

*b) Tactic analysis via tracking data*

For tactic analysis, Niu et al. [64] proposed a framework for systematically analyzing soccer tactics through the detection and tracking of field lines. The identified trajectory ultimately enables analyzing and improving soccer tactics. Comparably, the player trajectory was used for recognizing tactic patterns in basketball videos [65]. This type of framework was developed for directing focus to an open three-point attempt in a basketball game [66]. The trajectories of objects have frequently been used in event recognition [67], [68]. By employing a motion model for home and away team behaviors in soccer, Bialkowski et al. [69] visually summarized a soccer competition and provided indications of dominance and tactics. Recently, Zheng [70] conducted a survey on trajectory data mining, offering a thorough understanding of the field. More recently, a smart coaching assistant (SAETA) designed for professional volleyball training was introduced by Vales-Alonso et al. [71]. SAETA relies on a sensing infrastructure that monitors both players and their environment and provides real-time automatic feedback for aerobic and technical-tactical training in a team-sports environment.

*c) Tracking via camera networks*

Another critical objective is determining how to combine multiple cameras to obtain a more accurate tracking result. Choi et al. [72] reviewed problems regarding automatic initialization of player positions. Using SVMs, prior knowledge on the features of players can be retrieved, and depending on soccer match conditions, automatic initialization is often successful. However, reinitialization on tracking failure and guaranteeing a minimum time for initialization remain challenging. Yu et al. [73] presented an automatic camera calibration algorithm for broadcast tennis videos, evaluating problems regarding distortion, errors, and fluctuating camera parameters. The algorithm generates accurate camera matrices for each frame by processing the clips through of the following four steps: ground-feature extraction, camera matrix computation, camera matrix refinement, and clip-wise refinement. Figueroa et al. [74] used four cameras to observe the positions of all players on the soccer field at all times over the entire game. Their algorithm uses paths in a graph filled with blobs, representing segmented players, and analyzes different components of the blobs, such as the area and trajectory. Ren et al. [75] presented a technique for estimating the trajectory of a soccer ball by using multiple fixed cameras, despite constant movement and occlusion leading to size and shape discrepancies among the cameras over time. With a combination of motion information, expected appearance modeling, occlusion reasoning, and backtracking, the ball trajectory can be accurately tracked without velocity information.

*3) Naming Objects*

By taking advantage of state-of-the-art methods, detecting, tracking, and recognizing objects are feasible at present. However, the most difficult problem is identifying the detected object and the position of the athlete mentioned in subtitles. Most previous studies have focused on news videos because they are straightforward. For example, Satoh et al. [76] presented a system that identifies faces by associating the faces of the image with names in the captions. Everingham et al. [77] proposed a system that enables labeling the names of actors in TV episodes and movies. The framework integrates two procedures: 1) time-stamped character annotation by aligning subtitles and transcripts and 2) face tracking and speaker detection. An enhanced method was presented by Ramanathan et al. [78], who applied a bidirectional model to simultaneously assign names to the trajectories in the video and mentioned in the text. Normally, the naming process for news and movies videos is easier than that for sports videos, because stable face images and subtitles can be obtained easily. Lu et al. [79] used a conditional random field model to generate joint probability inferences regarding basketball player identities. A deformable part model detector was used to locate basketball players, and a Kalman-filter-based tracking-by-detection approach was employed to extract tracklets of players. Similarly, in [80], a large-scale spatiotemporal data analysis on role discovery and overall team formation for an entire season of soccer player tracking data is presented; the entropy of a set of player role distributions was minimized, and an EM approach was used to assign players to roles throughout a match. Their work can be applied future for identifying strategic patterns that teams exhibit. Nitta et al. [81] focused on identifying the structure of a sports game by using linguistic cues and domain knowledge to extract actors, actions, and events from a sports video and integrating the caption text and image stream. The system identifies actors who are performing an action and the action that they are performing.

However, face information is not observable at all times because the athletes usually run along an unpredictable path in three-dimensional space. An alternative approach for name assignment to objects is based on the text or numbers printed on athletes' clothes by using the video OCR [82] and scene text recognition techniques [83], [84]. To identify and track multiple players, Yamamoto et al. [85] proposed a tracking method that associates tracklets of identical players according to the results of player number recognition. Using a multicamera system, Pnevmatikakis et al. [86], [87] introduced a system that enables athlete tracking in different camera views and generates metadata such as locations and identities. Combining scene location, motion outliers, faces, and clothes enables tracking body hypotheses over time. For personalizing sports video broadcasts, Messelodi and Modena [88] presented an athlete identification module that applies scene text recognition to the embedded text from images. By applying their module to undirected and directed videos, precision rates of 98.9% and 88.7%, respectively, can be achieved. A method for soccer jersey number recognition by using a deep convolution neural network (CNN) was recently presented [89]. Two feature vector construction models, with the first treating every number as an individual class and the second treating each digit as a class, were employed and compared. With data augmentation and without dropout, the highest recognition rate achieved was 83%.

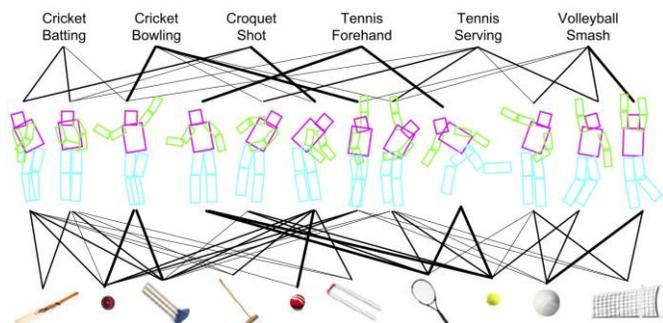

**Fig. 5**. The learned strength of connectivity between activities, human poses, and objects. The thicker lines indicate stronger connections, originally shown in [97].

*4) Action Recognition*

Action recognition has been extensively studied in recent years. Various approaches have been proposed for obtaining an accurate human body posture and movement estimate. Comparative explorations of recent developments in action recognition have described video-based [90]–[92] and image-based [93] approaches. However, only a few descriptions focus on the sports genre. Action recognition in sports can typically be classified into three categories: a) individuals, b) between objects, and c) between groups. The individual class requires an accurate human body posture with articulated parts and motion parameters for evaluating the global and local gesture performance. In the between-objects class, a player interacts with the competitor in winning the match or game. Here, the players hold a racket; thus, sports in this class are called racket sports. Sports in the between-groups class are called team sports. Almost all such sports are field sports. The members of a team work together to compete with the other team. Moreover, by combining the action class and field information, the event class can be determined.

Large-scale sports video analytics with deep learning is an emerging research area in many application domains. One of the most prevalent trends is to learn features by using a deep discriminatively trained neural network for action recognition. For example, Wang *et al.* [94] aggregated a hand-crafted feature [95] and feature determined through deep learning [96], using deep architectures to develop more effective descriptors.

*a) Posture and movement*

Regardless of the type of sport, the human body posture and movement of players have attracted much research attention. For example, Chen *et al.* [42] presented automatic generation of game information and video annotation through a pitching style recognition approach. In this approach, an object segmentation algorithm is first used to detect the pitcher's body. A star skeleton is then used to model the pitcher and recognize the pitching style. Yao and Li [97] applied a conditional random field model for action analysis. They modeled the mutual context of objects and human poses in human–object interaction activities. In their extended study [98], instead of modeling the human–object interactions for each activity individually, they ascertained overall relationships among different activities, objects, and human poses. As illustrated in **Fig. 5**, they modeled the *mutual context* between objects and human poses in human–object interaction activities so that one can facilitate the recognition of the other. Specifically, two pieces of contextual information are considered in the mutual context model. The *co-occurrence context* models the co-occurrence statistics between objects and specific types of human poses within each activity. The types of human poses, termed "atomic poses" (shown in the center of **Fig. 5**), can be considered a dictionary of human poses where the human poses represented by the same atomic pose correspond to similar body part configurations. In addition, they considered the *spatial context*, which models the spatial relationship between objects and different human body parts. Pecev *et al.* [99] shifted the focus from players to referees. Educational software that enables predicting the movement of basketball referees by using a multilayered perceptron neural network is beneficial in training young basketball referees.

Rather than depending on the vision-based method, Ghasemzadeh *et al.* [100] introduced a sensor-based method for tracking a baseball player's swing movements by using a wearable platform that yields multidimensional physiological data collected from a body sensor network. A semisupervised clustering technique is implemented to construct basic movement patterns, which can be used for detecting common mistakes and producing feedback. Schmidt [101] used detailed features such as angle displacements and velocities of kinematic chain articulations to analyze the movement patterns of free-throw shooters at different skill levels in a basketball video. In the experiments, several highly different movement organizations and their associated particular functionality were observed. The individual imprint on the movement pattern was shown to be related to its own, evidently appropriate functionality, which fit well with the personal constraints of the athlete in question.

Miyamori *et al.* [102] focused on a problem in which recognition easily becomes unstable because of partial tracking errors or lack of necessary information. They incorporated human behavior analysis and specific domain knowledge with conventional methods, developing an integrated reasoning module for robust recognition. The system was designed for tennis videos, and four typical video segments were used for reasoning object action: player behavior, ball position, player position, and curt-net lines. Kazemi *et al.* [103] addressed the problem of human pose estimation for football players, given images obtained from multiple calibrated cameras. With a body part-based appearance model, they used a random forest classifier to capture the variation in appearance of body parts in 2D images. The results of these 2D part detectors were then aggregated among views to produce consistent 3D hypotheses for the parts. The authors also presented a framework for 3D pictorial structures that can be used for multiple-view-articulated pose estimation [104]. Swears *et al.* [105] proposed a modeling approach, called Granger constraints DBN, which enables modeling interactions between multiple co-occurring objects with complex activity. To simplify the measurement of temporal dependence, they used the Adaboost feature selection scheme to discriminatively constrain the temporal links of a DBN.

TABLE II
BENCHMARK DATA SETS AVAILABLE FOR SPORTS VIDEO ANALYSIS; IN THE "TASKS" COLUMN, "$A_R$," "$A_L$," "$P$," AND "$C$" RESPECTIVELY REPRESENT "ACTION RECOGNITION," "ACTION LOCALIZATION," "POSE ESTIMATION," AND "SPORTS TYPE CLASSIFICATION"

| Systems | Dataset(year) | #sports/actions categories | Source | #instances | Tasks | Locations |
|---|---|---|---|---|---|---|
| [136] | Wang et al. (2006) | 3[a]/25 | Internet | 1400, 4500, 8500 images | $A_R$ | [137] |
| [126] | UCF Sports (2008) | 10[*] | Broadcast TV (e.g. ESPN and BBC) | 15 sequences. | $A_R, A_L$ | [127] |
| [141] | Olympic Sports (2010) | 16[*] | YouTube | 50 videos | $A_R$ | [142] |
| [132] | ACASVA (2011) | 2[b]/3 | Broadcast TV | 6 sequences | $A_R, A_L$ | [134] |
| [103] | KTH (2013) | 1[c,*] | Staged | 3D: 2400 images 2D: 5907 images; | $P$ | [135] |
| [138] | Sport-1M (2014) | 487[*] | YouTube | ~1.1 million videos | $C$ | [139] |
| [143] | SVW (2015) | 44/30 | Smartphone & Tablet | 4200 videos | $A_R, A_L$ | [144] |

[a]Figure skating, baseball, and basketball
[b]Tennis and badminton
[c]Football
[*]No significant distinctions between sports types and actions categories

*b) Feature representation*

Regarding action recognition, two critical learning phases are highly required: feature representation and classifier learning. Feature representation extracts and ranks useful 2D or 3D features such as Gabor features [106], histograms of oriented gradients (HOGs) [107], gradient location-orientation histograms [108], motion history images [109], motion history volumes [110], 3D gradients [111], and other volumetric (voxel) data [112]–[115]. The common strategy is based on a predefined articulate model for matching human body parts [116]–[118].

*c) Classifier learning*

When feature vectors are extracted, a classification framework is used to recognize the types of actions. For instance, Zhang et al. [119] proposed an effective action recognition method based on the recently proposed overcomplete independent components analysis (ICA) model. Instead of using a pretrained classifier, they adopted the response properties of overcomplete ICA to perform classification. In this approach, a set of overcomplete ICA basis functions is learned from 3D patches from training videos for each action. The test video can be labeled as the action class whose basis functions can reconstruct the video with the smallest error. Similar to Zhang's method, Le et al. [120] presented a hierarchical invariant spatiotemporal feature learning framework based on independent subspace analysis. For modeling movement patterns, a structured codebook-based method with an SVM classifier was proposed in 2014 [121]. For modeling the temporal transition of postures, Li et al. presented a continuous HMM with a left–right topology [122]. In addition, Taylor et al. [123] extended the gated restricted Boltzmann machine (GRBM) [124] from 2D images to 3D videos for learning spatiotemporal features and named the extension the convolutional GRBM method.

*d) Data sets*

Unlike [125], the current survey focused on reporting sports action data sets that do not explicitly consider comprehensive action data sets. Here, we reported seven sports action data sets. Two of these data sets presented the image source with action markers, and the remaining five comprised a collection of video clips. A comprehensive list of action datasets for sports with corresponding details is provided in **Table II**.

The UCF sports data set [126] contains 150 video sequences with a resolution of $720 \times 480$ concerning 10 disciplines, namely diving, golf swinging, kicking, lifting, horseback riding, running, skating, baseball swinging (split into around high bars and on the pommel or floor), and walking. All video clips were collected from sports broadcasts by networks such as ESPN and the BBC. The data set exhibits high intraclass similarity between videos and large variation in interclass viewpoint changes and noise. It was published by Rodriguez et al. and is available on the Internet [127]. Since the release of this data set in 2008, the average recognition rate has increased, being 85.6% in 2009 [128], 87.3% in 2010 [129], 86.8% in 2011 [130], 95% in 2012 (used extra training data) [131], and 89.7% in 2014 [121]. This data set is the most widely used among the six sports action data sets.

A tennis data set called ACASVA, published by de Campos et al. [132], was used to compare the BoW and space-time-shape approaches for action recognition in videos. The tennis game videos were reproduced from TV broadcasts in standard resolution, and Yan et al.'s ball tracker [60] was used to detect relevant instances. The data set and MATLAB software package [133] are available in [134].

The KTH Multiview Football data set [103] is a 3D data set containing 2400 images of football players obtained from three views at 800 time frames with 14 annotated body joints and 5907 images with the annotated 2D pose of the player. The data set is available in [135].

Wang et al. [136] presented a data set containing 10 labels in figure skating clusters, seven labels in baseball clusters, and eight labels in basketball clusters. The dataset can be acquired from [137].

Karpathy et al. [138] constructed an immense data set named Sports-1M, which consists of greater than 1.1 million annotated YouTube videos for 487 sports. The data set is available in the GitHub repository [139]. Compared with the UCF sports dataset, Sports-1M entails greater difficulty in feature recognition. The highest recognition rate is approximately 74.4% with a two-stream CNN and temporal feature pooling [140].

The fifth set is an Olympic sports data set [141] containing 50 videos from each of the following 16 classes: high jump, long jump, triple jump, pole vault, discus throw, hammer throw, javelin throw, shot put, basketball layup, bowling, tennis serve, diving (split into platform and springboard), weightlifting (split into snatch and clean and jerk), and vault (gymnastics). This data set is available in [142].

A more recently introduced data set, sports videos in the wild (SVW) [143], contains 4200 videos covering 30 sports categories and 44 actions. The videos in this data set were captured by users of the leading sports training mobile app Coach's Eye while practicing a sport or watching a game. This data set is available in [144].

TABLE III
COMPARISONS OF THE WORK BEING DONE IN THE DOMAIN OF SPORTS EVENT ANALYSIS

| Systems | Data sources* | Methodologies | Sport genres | Event examples | Application views |
|---|---|---|---|---|---|
| [69] | m-visual | Expectation-Maximization (EM) method, Hungarian algorithm | Soccer | Possession condition, attacking and defending deep | Possession analysis, game formation analysis, tactic analysis |
| [145] | Visual | A priori model comprises four major components: a soccer court, a ball, players, and motion vectors. | Soccer | Kick and goal at different place | Content-based video retrieval |
| [146] | Audio, visual | Grass Coloured Pixel Ratio Characteristic (GCPRC) | Gaelic Football | Goals-points | Event detection |
| [147] | Semantic | Color-based shot boundary algorithm | baseball | Outs, scores, base-occupation | Summary, indexing for sports video |
| [148] | Visual | Hidden Markov model (HMM), probabilistic model | Baseball | Home run, catch, hit, infield play | Highlight detection and classification |
| [149] | Visual, audio | Music Analysis Retrieval and Synthesis for Audio Signals (MARSYAS), Support Vector Machine (SVM) | Tennis, soccer | Goals | Summarization, automatic annotation |
| [150] | Audio, visual | SVM | Soccer, rugby, hockey, Gaelic football | Goals, play, break | Event detection, shot boundary detection |
| [151] | Visual | Entropy-based Motion Analysis, homoscedastic error model | Basketball, soccer, tennis | Basketball : scoring, miss, free throw Soccer : goal, free kick, corner Tennis : fault, ace, volley | Video segmentation, event detection |
| [152] | Semantic, l-visual | Global Motion Estimation (GME), RBF neural networks and decision-tree classifier, predefined finite-state machine model | Non-specified | High jump, long jump, javelin, weight, throwing, and dash | Events recognition in sports video |
| [153] | l-visual | HMM | American football, baseball, sumo wrestling | Play-break | Event detection, video summary |
| [154] | Visual | HMM | Soccer | Goal, corner kick | Video browsing, events detection |
| [155] | Audio, visual | Binary logo model | Soccer, Australian Football League(AFL), basketball | Soccer: goal, shoot, foul AFL: goal, behind, mark, tackle Basketball: goal, free throw, foul | Content-based video retrieval, event detection |
| [156] | Audio, m-visual | SVM, Radial Basis Function (RBF), | Tennis, table tennis | Cheer, silence, excited speech | Summarization, event detection on racquet sports video |
| [157] | Text, m-visual | Connected Component Analysis (CCA) | Basketball | Score change | Real-time score detection |
| [158] | Audio, visual, text | Temporal model | Basketball | Goals | Content-based video retrieval, sports video analysis |
| [159] | Visual | Bayesian Belief Network (BBN), SVM | Baseball | Strike out, homerun, sacrifice fly, hit, hit and run, fly out or ground out, double play, walk, steal base, wild pitch, passed by or balk, others | Scoreboard recognition, shot transition classifier, scoreboard identification |
| [160] | Audio | SVM | baseball | Hit the ball, exciting speech | Highlight extraction, summarization |
| [161] | Audio | Entropic Prior Hidden Markov Models (EP-HMM) | Baseball, golf, soccer | Applause, cheering, music, speech | Highlight extraction |
| [165] | Audio, visual, text | Dynamic Bayesian Networks (DBNs) | Formula 1 races | start, fly out, passing | Highlight detection |
| [166] | Semantic | DBN, temporal intervening network | Soccer | Goal, corner kick, penalty kick, card | Highlight extraction |
| [167] | Audio, visual | Maximum entropy model (MEN) | Baseball | Home run, outfield hit, outfield out, infield hit, infield out, strike out, walk | Highlight detection, machine learning, video content analysis |
| [162] | Visual | Multi-level Semantic Network (MSN), BBN | Baseball | Highlights | Semantic feature extraction, highlight detection |
| [168] | Visual, semantic | Hidden Markov Model under normal (NHMM) and enhanced (EHMM) observations | Soccer | Placed, foul, shoot, goal, highlight | Video retrieval and personalized sports video browsing, shot classification |
| [169] | Audio, text | Recognition of the textual overlays | American football | Touchdown, field goal | Personalized video abstraction, highlights detection |
| [170] | Visual | Model highlights using finite state machines | Soccer | Forward pass, shot on goal, turnover, placed kick, corner, free kick, penalty, kick off, counterattack | Semantic annotation, highlights detection |
| [171] | Visual | Camera motion parameters | American football | Short and long pass, short and long running, quarterback sacks, punt, kick off | Event detection |
| [172] | l-visual | SVM, Camera label estimation | American football | Goal, foul, corner, substitution | Event tagging, spatial segmentation |
| [174] | Visual | Physics-based method of ball trajectory extraction | Volleyball | Serve, Receive, Spike | Ball tracking, action detection |
| [176] | Visual | Switching Probabilistic Principal Component Analysis (SPPCA), boosted particle filter (BPF), Histograms of Oriented Gradients (HOG), Sparse Multinomial Logistic Regression (SMLR) | hockey | N/A | Objects tracking, action recognition |
| [177] | l-visual | HMM | Basketball, football, racing, boxing, soccer | Commercial, slow motions | Slow motion replay segments detection, Video summarization highlights generation |
| [178] | Audio, visual, temporal | Subspace-based data mining | Soccer | Corner, goal | Semantic event detection, concept detection |
| [179] | Audio, visual, semantic, attention | Support Vector Regression (SVR) | Tennis, badminton | Overhead-swing, left-swing, right-swing | Affective analysis, highlight ranking, video browsing, action recognition |
| [184] | Visual, text, semantic | Combination of domain-independent global model-based filtering methods with domain-specific object-level spatiotemporal constraints | Tennis, basketball | N/A | Summarization and browsing |
| [187] | l-visual | Decision-tree learning method, Entropy-based inductive tree-learning algorithm | Basketball | Left court, right court, middle court, other | Video indexing, classification, summarization |
| [188] | l-visual | Advanced temporal pattern analysis and multimodal data mining method, decision tree learning algorithm | Soccer | Goal | Event detection, summarization, browsing and retrieval |
| [189] | Visual | Probabilistic Support Vector Classification (PSVC) | Soccer | Goal, shot, corner, free kick | Object tracking, tactic analysis |
| [190] | Visual | Circle Hough Transform (CHT) | Soccer | Goal | Real-time goal detection |
| [191] | Visual, text, semantic | Temporal Neighboring Pattern Similarity (TNPS) measure, Probabilistic Latent Semantic Analysis (PLSA), Conditional Random Field Model (CRFM). | Soccer | Goal | Summarization, event detection, provision of augmented information |
| [192] | Visual, text, audio, semantic | Time Interval Maximum Entropy (TIME) | Soccer | Goal, yellow card, substitution | Event detection in video documents, indexing |
| [193] | Visual, text | HMM | Soccer | Goal, red card, yellow card, penalty, game start, game end, half time, disallowed goals | Summarization, event detection |
| [194] | Visual, semantic | Closed caption (CC) text | American football | Touchdown, field goal | Event-based video indexing |

*l-visual: low-level visual features; m-visual: mid-level visual features

### B. Highlight Detection and Event Recognition

Because an increasing amount of research is focused on large-scale content-based multimedia mining, there is a lack of systematic surveys specifically designed for sports videos. This section mainly reviews the event-oriented state-of-the-art studies according to the type of data source, methodology, sports genre, desired event, and application field. The reviewed articles are listed in **Table III**.

#### 1) Scene Change Detection

An event is composed of a group of video scenes or shots. The shot boundary detection technique aims to divide a long video sequence into video segments. Gong *et al.* [145] proposed temporal segmentation for partitioning a video sequence into camera shots. Each shot contains identical semantic concepts. Several studies have used scene change detection to perform event classification. The simplest approach to detecting transitions of shots is seeking discontinuities in the visual content of a video sequence. For instance, the authors in [146]–[149] have proposed color-based shot boundary detection schemes for segmenting a video. Poppe *et al.* [149] used color histogram differences for detecting scene changes. During the broadcast of baseball games, multiple broadcast cameras were mounted at fixed locations in the stadium. The authors first built statistical models for each type of scene shot with histogram products [148]. In addition, Sadlier and O'Connor [150] employed their own algorithm to solve the problem caused by the high-tempo nature of field sports; during live action segments, the broadcast director has little opportunity to use shot transition types other than hard shot cuts. In [151], an event-based segmentation method was presented and a motion entropy criterion was employed to characterize the level of intensity of relevant object motion in individual frames. Because global motion usually changes largely in shot boundaries, Wu *et al.* [152] detected abrupt augmentations of the magnitude of accelerations in the frame sequence and marked them as semantic clip boundaries.

#### 2) Play-and-break Analysis

Designing a generic model for analyzing all types of sports videos is difficult, because the event is a domain-dependent concept. Nevertheless, a compromised approach, called play-and-break analysis, has been presented. This analysis roughly models sports games. Li and Sezan [153] used both deterministic and probabilistic approaches to detect critical events, called plays. An iterative algorithm was used to detect plays until a nonplay shot was identified. On the basis of this play information, games could be summarized. Xie *et al.* [154] applied statistical techniques for analyzing the structure of soccer videos. Plays and breaks (i.e., nonplay) are two mutually exclusive states of soccer videos. First, experimental observations were used to create a salient feature set. HMMs and dynamic programming algorithms were then applied for classification and segmentation. Tjondronegoro and Chen [155] pioneered the use of a play–break segment as a universal scope of detection and a standard set of features that can be applied to different sports. Specially, a racket sports game consists of many play and break events. In [156], the play event in the racket sports, which has a unique structure characteristic, was referred to as "rally." That study segmented a racket sports video into rally and break events and ranked the rally events according to their degrees of excitement.

#### 3) Keyframe Determination

Keyframe detection is a crucial operation performed after the retrieval of shot boundaries. A keyframe is used to represent the status of each segment [145], [147]. Liang *et al.* [147] proposed a framework for detecting data changes in superimposed captions and employing rule-based decisions to detect meaningful events in a baseball video. They considered only the number of outs, number of scores, and base-occupation situation, and thus, they could simply select any frame as the representative shot (segment) if the caption information remained unchanged. In addition, numerous studies have focused on scoreboard detection [157]–[159]. First, a scoreboard subimage was located using the difference in consecutive frames and the gradient information of each frame. Subsequently, the changed pixels in the scoreboard subimage were evaluated [157]. Nepal *et al.* [158] used the displayed scoreboard as the second key event, and developed temporal models based on the pattern of occurrence of the key events observed during the course of the game. Regarding the structure of a baseball video, the critical events conventionally occur between pitch scene blocks [147], [153], [159].

#### 4) Highlight Detection

Highlights are critical events in a sports game that must be extracted. Highlight extraction requires specific features and analyzers. A video consists of audio and visual features. Several previously proposed frameworks have applied audio features instead of classifiers, such as SVMs [149], [156], [160], HMMs [161], Bayesian belief network (BBNs) [162]–[164], DBNs [165], [166], maximum entropy models [167], and multilevel semantic networks, which are an extension of the BBN. For instance, Petkovic et al. [165] proposed a robust audiovisual feature extraction scheme and a text detection method, and used classifiers for integrating multimodal evidence from different media sources to determine the highlight events of a Formula 1 race.

TABLE IV
VISUAL FEATURES FOR CONTENT ANALYSIS

| Visual features | References |
|---|---|
| Captions and text | [158], [196], [201], [208], [212], [221], [222], [223] |
| Playfields | [145], [166], [199], [203], [204], [208], [211], [216] |
| Camera motion | [155], [162], [166], [171], [172] |
| Ball positions | [170], [173], [174], [190] |
| Player positions | [148], [166], [170], [173], [174], [176], [179], [187] |
| Replays | [149], [155], [166], [169], [177] |

In sports, visual features for highlight detection can be classified into six groups: 1) captions and text, 2) playfields, 3) camera motion, 4) ball positions, 5) player positions, and 6) replays. First, an algorithm addressing text detection [168] and scoreboard recognition was presented in [169], which employed an overlay model to identify the event frame. Second, given a "playfield" frame, the region of the image that represents the playfield was roughly identified from grass color information by using color histogram analysis [170]. Unlike grass field regions, audience regions have different textures. Therefore, Chao et al. [166] used edge density information to detect audience regions. Third, Lazarescu et al. [171] used camera motion parameters, including the number of stages in camera motion, tilt motion, and camera angle, in a method for detecting seven types of American football plays. A more recently presented system [172] identifies the type of camera view for each frame in a video. The system was designed for football games, and salient events such as goals, fouls, corners, substitutions can be distinctly tagged on the timeline.

Generally speaking, balls and players play the main roles in a sports competition. Thus, ball and player tracking techniques have typical features in sports video analysis. Kobayashi et al. [173] input vector time-series data consisting of player, referee, and ball positions into a data-driven stick breaking hidden Markov model for predicting key activities and dominance conditions in a soccer game. In addition, Chen et al. [174] addressed the obstacles and complexities involved in volleyball video analysis by developing a physics-based scheme that uses motion characteristics to extract the ball trajectory, despite the presence of numerous moving objects. In [175], temporal and spatiotemporal regions of interest were used to detect the scoring event and scoring attempts in basketball mobile videos. Furthermore, Lu et al. [176] used a panning, titling, and zooming camera to automatically track multiple players while detecting their action. The system involves using HOGs, an boosted particle filter, and the combination of HOG descriptions with pure multiclass sparse classifiers. Finally, slow motion replays were modeled using an HMM, which emphasizes highlights and critical events [177]. These visual cues are summarized in **Table IV**. In addition, Shyu et al. [178] proposed a subspace-based multimedia data mining scheme for event detection and used a decision tree for training event detectors. Zhu et al. [179] presented a multimodal approach for organizing racket sports video highlights through human behavior analysis. Support vector regression constructs a nonlinear highlight ranking model. Furthermore, some previous studies have focused on not only highlight extraction but also highlight ranking [156], [169].

5) *Use of Domain Knowledge*

Sports events are typically composed of several video shots that appear in a certain temporal order. By using shot boundary detection techniques, the video can be segmented into scenes. Domain knowledge (i.e., paradigm) can be used to determine the type of video shot. Xie et al. [154] employed domain knowledge to perform event recognition. Huang et al. [180] proposed a temporal intervening network for modeling the temporal actions of particular events in a soccer game, such as goal, card, and penalty events. In [181], a Bayesian network-based model was used for event detection and summarization applications in a soccer video, whose structure was estimated using the Chow-Liu tree [182], and the joint distributions of random variables were modeled using the Clayton copula [183].

Temporal video patterns typically vary depending on the type of sport. In other words, sports video analysis is a domain-specific problem. Several extant systems analyze only sports that involve an identical game structure [15], [20], such as court-net sports including tennis, badminton, and volleyball, to obtain a comprised result. Many state-of-the-art frameworks have added domain knowledge to increase the accuracy. For example, Zhong et al. [184] presented a generic domain-independent model with spatiotemporal properties of segmented regions for identifying high-level events, such as strokes, net plays, and baseline plays. Similarly, analyzing the temporal relationships among video shots enables automatic headline detection in TV sports news [185]. Chen et al. [186] introduced an HMM-based ball hitting event exploration system for baseball videos. A rule-based model might require domain knowledge and employ the decision-tree learning method to calculate rules and apply them to low-level image features [187]. The investigation of decision tree algorithms and multimodal data mining schemes was the main contribution of [188].

Zhu et al. [189] focused on tactical analysis, extracting tactical information from attack events in broadcast soccer videos to assist coaches and professionals. In soccer, a goal event must be preceded by the offense events that were considered in [190] and [191]. D'Orazio et al. [190] used four high frame rate cameras placed on the two sides of the goal lines. Snoek et al. [192] presented the time interval maximum entropy (TIME) framework for combatting multimodal indexing problems regarding representation, inclusion of contextual information, and synchronization of the heterogeneous information sources involved. The TIME framework proved to be successful, evidently reducing the viewing time of soccer videos. Regarding application, Nichols et al. [193] attempted to summarize events by using only Twitter status updates. In addition, Babaguchi et al. [194] proposed an event-based video indexing method that involves using time spans where events are likely to occur and keywords from the closed caption stream. Chen et al. [188] emphasized the framework of mining event patterns. Using a group of audiovisual features and domain knowledge, they applied temporal pattern analysis for semantic event mining. Gupta *et al.* [195] applied an AND-OR graph as a representation mechanism for storyline models and learned the structure and parameters of the graph from roughly labeled videos by using linguistic annotations and visual data. They showed a storyline through the instantiation of AND-OR graph for a baseball game broadcast from news videos.

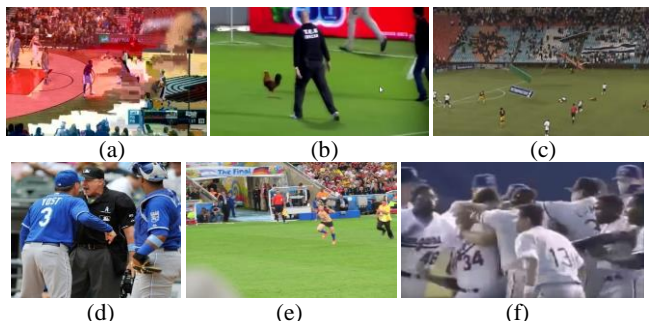

**Fig. 6**. Examples of unusual situation of the game.

The temporal patterns of video shots reveal critical clues for accessing video content. Event recognition can be directly achieved using a temporal pattern. For instance, the goal event in soccer, comprising the occurrences of gate appearance, a close-up view, and replays, follow certain rules of causality. After the gate appears, the mid-distance view is shown for a short duration. The scorer celebrates with the other teammates. Afterward, the scene transitions rapidly and the first replay video is shot in slow motion. Other replay shots with different views, as well as scenes of the gate and net, follow. Finally, the scoreboard appears, the goal event concludes, and the bird's eye view of the entire field signals continuation of the match. If audio information is supplied, the loud cheering sound from the audience can serve as a useful clue for detecting the occurrence of a goal event.

For a home-run event in a baseball game, the scene begins with a typical moment before the pitch. The players are positioned in the designated locations, and thus, this scene is simple to identify by matching the color layout. If an audio signal is detected, the announcer sounds surprised after the pitch. To track the ball, the scene moves very rapidly. If the remaining time is sufficient, the auditorium is presented. Half of the audience and half of the fielder who is the closest to the outfield wall are captured, and the batter running the bases and the pitcher are shown in close-ups. Subsequently, a cheering fan is inserted into the frame. Next, several replay shots in slow motion at different angles are captured. Sometimes, the hit is caught by a high-speed camera. After the replay, the camera shoots the player who returns to home plate and captures a close-up shot of teammates clapping. When the next batter walks toward home plate, the video frame returns to the initial scene.

In a basketball game, there are fewer regular temporal patterns for event classification. The scoreboard is conventionally used to determine whether the shooter scores a two- or three-point shot. When the scoreboard changes, a close-up of the scorer is shown. Offense and defense exchange and pass the ball from the baseline. Events can be identified by players' trajectories and the path of ball passing. For example, pick and roll is a frequently used tactic in basketball games. During the game, the offense is arranged in pairs with the defense. When a pick-and-roll event occurs, the ball keeper faces the defense first. Another offensive player moves close to the ball keeper, obstructing the defense's path. Players' movement paths become discontinuous, and only the ball keeper can move freely. Finally, the ball keeper suddenly moves to the free-throw lane and passes the ball to the shooter, and shooter shoots the ball toward the basket. Accordingly, to determine the event category in more detail, the object trajectory must be tracked precisely and motion analysis must be performed. The framework of 3D camera modeling has also been applied for event classification efficiently. Su *et al.* [196] upgraded the feature extraction part for event classification from the image domain to the real-world domain. The real-world positions of objects (i.e., athletes) could be determined using an automatic 3D camera calibration method. At the scene level, combining a Bayesian modeling approach with player segmentation and tracking and playing-field knowledge, the system could classify events such as service, net approaches, and baseline rallies.

In summary, domain knowledge enables a reasoning system to obtain a compromised result. However, uncertainty occurring in the playfield, such as a moving bird, a strong wind, and transcoding problems, cannot be avoided, as shown in **Fig. 6(a)–(c)**. Unpredictable interruptions caused by streakers, arguments between the referee and coach, and players fighting are shown in **Fig. 6(d)–(b)**. These extraordinary activities are rare. Nevertheless, context-based semantic analysis can be adopted to assess unusual events.

### C. Contextual Inference and Semantic Analysis

In sports, the context is composed of event outcomes and match summaries. The semantic meanings of the context are critical for the game. The context can typically be obtained from captions or visual cues. Comparisons of recent context-oriented studies are shown in **Table V**.

*1) Context Extraction*

***Context from captions:*** In a broadcast sports video, a superimposed caption box (SCB) embedded in the frame is a typical means for quickly conveying the ongoing game status to viewers. To detect the SCB and read captions, Guo *et al.* [197] applied SIFT point matching and Tang *et al.* [198] used a fuzzy-clustering neural network classifier. Moreover, Su *et al.* [196] presented a caption model for identifying the meanings of the captions. Zhang *et al.* [199] used a set of modules including caption box localization, segmentation, and a Zernike feature combined with temporal transitional graph models to recognize superimposed text. Sung *et al.* [200] developed a knowledge-based numeric caption recognition system to recover valuable information from an enhanced binary image by using a multilayer perceptron neural network. Overall, most caption recognition systems focus on recognizing the text through optical character recognition [201], [202] and employing a template matching algorithm [203], [204] to confirm the caption according to either numbers or text. However, recognizing symbol captions remains challenging, and thus, caption interpretation has emerged as a critical research topic. Lie *et al.* [205] applied visual feature analysis to

TABLE V
COMPARISONS OF CONTEXT-ORIENTED APPROACHES

| Systems | Data source* | Analyzing model | Sport genres | Context examples | Application view |
|---|---|---|---|---|---|
| [196] | Temporal, l-visual | Model-based segmentation approach, OCR | Baseball, Basketball, Rugby, Soccer | N/A | Sports captions segmentation |
| [200] | Text | Multiplayer perceptron (MLP) network, Error back-propagation (BP) algorithm | Baseball | Numeric captions | Caption recognition |
| [205] | Temporal, Text, l-visual | Algorithm integrating caption rule-inference, Visual feature analysis | Baseball | Strikes, Balls, Outs, Scores, Base-occupies | Baseball video retrieval and summarization |
| [206] | Temporal, Text, l-visual | Superimposed caption box(SCB) color model, Probabilistic labeling algorithm | Basketball, Baseball | Inning, Scoring, Ball, Strike Out, The name of the team, Quarters | Caption template extraction and identification |
| [207] | Temporal, Text, l-visual | SCB color model, Connected Component Analysis (CCA), SCB interpretation | Basketball, Baseball | Quarters, The score of the home team, The score of the visiting team | Caption template extraction and identification |
| [225] | Temporal, audio Semantic, Text | Latent semantic index based text event detection, naïve Bayesian network | Soccer | Goal, Shot, Corner, Free Kick, Card, Foul, Offside, Substitute | Event Annotation |
| [226] | Audio, Text, l-visual | Gray level co-occurrence Matrix (GLCM) | Basketball, Soccer | Goal, Penalty, Red card, Shot on Goal, Attempt, Free Kick, Yellow card, Offside, Foul, Corner, Substitution, 3-point shot, 2-point shot, Free Throw, Rebound, Turnover | Video summarization, Event annotation |
| [227] | Temporal, Text | Hidden Markov Model(HMM) | Basketball | Shot, Jumper, Lay-up, Dunk, Block, Rebound, Foul, Free throw, Substitution | Events annotation and indexing in the video |
| [229] | Text | Hierarchical keywords search | Basketball, Soccer | Corner, Shot, Foul, Card, Free kick, Offside, Substitution, Goal, Jumper, Layup, Dunk, Block, Rebound, Free throw | Video annotation, retrieval, Indexing and summarization |
| [230] | Temporal, Text | Conditional random field model (CRFM), Expectation Maximization (EM) | Basketball, Soccer | Shot, Jumper, Layup, Dunk, Block, Rebound, Foul, Free throw, Substitution, Corner, Card, Free kick, Offside, Goal | Sports video summarization and retrieval |
| [231] | Audio, Temporal, Text | Basic Criterion, Greedy Criterion, Play-Cut Criterion | Baseball | Homerun, Sacrifice-hit, Two-base-hit, Single-hit, Strike-out… | Video abstraction |

*l-visual: low-level visual features

develop a system that can classify baseball events into 11 semantic categories, which include hitting and nonhitting as well as infield and outfield events. Shih *et al.* [206] performed caption interpretation to analyze the SCB in sports videos. They divided the captions into three categories: proportion, inverse, and specific types. Similarly, Shih and Huang [207] presented a method that interprets the SCB where the SCB template cannot be determined *a priori*. A framework designed for golf videos was proposed in [208].

*Context from visual cues:* Contextual annotation validates essential insight into the video content while providing access to its semantic meanings. For sports video analysis, the extraction of external metadata from closed captions have been employed in many applications such as video indexing, retrieval, and summarization. In comparison with visual features, contextual information presents a semantic inference regarding the video content. Steinmetz conducted comprehensive research by using a context model to perform semantic analysis of video metadata [209]. The research involved multiple knowledge bases. In the semantic inference process, the context is produced dynamically, considering the characteristics of the video metadata and the confidence values. The research findings can be applied in analyzing textual information originating from different sources and verifying different characteristics. Growing evidence indicates that, by examining the semantic context, video analysis can be effectively used to improve content-based video processing procedures such as keyframe extraction [210], [211], sports video resizing [212], concept-based video indexing [213], and video search [214], [215]. In addition, a mapping mechanism between the context and content designed for baseball videos was presented by Chiu *et al.* [216]. The mechanism focuses on aligning the webcast text and video content. They proposed an unsupervised clustering method called hierarchical agglomerative clustering for detecting the pitch segment in baseball videos. In addition, they applied a modified genetic algorithm to align the context of webcast text and video content.

Most notably, the trend for reading text in images and videos has shifted to the use of advanced learning techniques. For example, Jaderberg *et al.* [217] presented a word detection and recognition system and used a multiway classification model with a deep CNN. Li *et al.* [218] employed a fine-tuned deep CNN to extract visual features, and fed these features into recurrent neural networks (RNNs) to generate representative descriptions for video clips. Similarly, Venugopalan *et al.* [219] proposed a video-to-text generator that uses deep RNNs for the entire pipeline from pixels to sentences.

*2) Text Analytics*

For text analysis, Chen *et al.* [220] used the AdaBoost algorithm to generate a strong classifier and detect texts. Lyu *et al.* [221] presented a multilingual video text detection, localization, and extraction method, specifically for English and Chinese. Xi *et al.* [222] proposed a text information extraction system for news videos. For detecting and tracking the text, multiframe averaging and binarization were applied for recognition. Lienhart *et al.* [223] presented a method for text localization and segmentation for images and videos, and for extracting information used for semantic indexing. Noll *et al.* [224] revealed the effective combination of local (the measures are derived from the local attributes of the features, e.g., the angles of a corner) and context similarities for object recognition. A Hough-based matching algorithm was introduced for analyzing corner feature similarities. More recently, Wang *et al.* [225] transformed an event detection problem into a synchronization problem. They annotated soccer video events by using match reports with inexact timestamps. The game start time was first detected for a crawled match report, and the naïve Bayesian classifier was applied to identify the optimal match event.

*3) Semantic Event Extraction*

For semantic event extraction, conventionally, parse trees are first used to extract semantic events and then hierarchically arranged to provide input for a logic processing engine to generate a summary [226]. Zhang *et al.* [227] employed a multimodal framework to extract semantic events in basketball videos. They combined text analysis, video analysis, and text or video alignment for semantics extraction, event moment detection, and event boundary detection. Zhang and Chang [228] developed a system for baseball video event detection and summarization by using superimposed caption text detection and recognition. More recently, Chen *et al.* [229] extracted semantic events from sports webcast text by using an unsupervised scheme. A filtering technique is implemented to remove unrelated words, and the remaining words are sorted into categories where keyword extraction is executed to recognize crucial text events. Similarly, a system capable of analyzing and aligning webcast text and broadcast video for semantic event detection was introduced for the automatic clustering of text events and extraction of keywords from webcast text [230]. Nitta *et al.* [231] proposed a method for automatically generating both dynamic and static video abstracts from broadcast sports videos. The method relies on metadata for semantic content and player influence and user preferences for personalization.

## IV. CHALLENGES AND FUTURE DIRECTIONS

In the past decade, many research achievements in sports video content analysis have been witnessed and have highly influenced the direction of prospective research. In this section, we highlight the prominent challenges identified in the literature and categorize the potential directions in sports video content analysis for future research into three classes: 1) sports video content analysis in the media cloud, 2) large-scale learning techniques for deeper context discovery, and 3) robust visual feature extraction for precise human action recognition.

*A. Sport Video Content Analysis in Media Cloud*

Recently, many portable devices for video recording, such as the camera handset, personal video recorder, and tablet, have been developed. Consequently, a device user can be a content producer. Users are interested in sharing their videos through social networks, and this has become a cultural phenomenon. Currently, the focus of major research efforts in sports video content analysis such as sports type classification [33] and salient event detection [175] has shifted to mobile videos. According to the white paper Cisco Visual Networking Index (VNI): Forecast and Methodology, 2014–2019 [232], content delivery networks will carry 62% of Internet traffic worldwide by 2019. The number of devices connected to Internet Protocol networks will be three times as high as the global population in 2019. Therefore, research on video caching, transcoding, and adaptive delivery in the media cloud is urgently required [233]. The major challenge is minimizing the overall operational cost regarding the usage of buffering, computing, and bandwidth resources. For example, a mobile video–audio mash-up system called MoVieUp [234] was presented for collating videos shared by client users. It operates in the cloud to aggregate recordings captured by multiple devices from different view angles and different time slices into a single montage. This system was designated for concert recordings. Sports videos do not have a continuous and smooth audio stream for stitching together a video–audio mash-up. In sports applications, the mobile devices normally serve as transceivers to avoid tasks with high computational cost. How to combine sports recordings captured by multiple mobile devices remains an open question. The trend of crowd sourcing raises substantial challenges to cloud service providers. In the future, a synthetic 3D montage of the current sports field could be constructed using breakthrough object and scene alignment and synthesis techniques.

*Scalability* is among the most crucial challenges for cloud mobile media networks [235]. Previous studies have generally addressed the scalability in temporal motion smoothness and spatial resolution instead of content-aware scalability. Client users demand to receive the status of a desired sports competition. Fine-grained video content transcoding is configured under different channel conditions associated with the bandwidth resources and end devices. The client users receive various details on the sports content at varying degrees of precision such as score statistics, audio broadcasts, player portraits, highlight images, an edited video segment, and an entire video broadcast. Another future objective is to combine semantic web tools and technologies [236] to overcome the aforementioned challenges [237].

*B. Large-scale Machine Learning Techniques for Deeper Information Discovery*

With the proliferation of media devices, the media content that we access conceals extremely valuable information and knowledge. The era of big data is considerably changing what we know and how we know it in the information world. Big data analytics enables discovering the latent semantic concepts embedded in video content. Sports data analysis is becoming large scale and diversified. Advanced developments in machine learning lead to solutions to real-world large-scale problems.

Since 2006, with the surge of deep learning, research on visual understanding under the paradigm of data-driven learning reached a new height [238]. Deep learning methods such as CNNs, restricted Boltzmann machines, autoencoder and sparse coding have been adopted and demonstrated to be successful for various computer vision applications. One of the most salient achievements is action and activity recognition through deep learning, which has been performed widely in recent years [94], [96]. Several studies focused on sports video analysis have employed deep learning. For example, CNNs have demonstrated superiority in modeling high-level visual semantics [89], [217], and recurrent neural networks have exhibited promise in modeling temporal dynamics in videos [218], [219].

Nevertheless, there is substantial room for development in sports video analysis with deep learning. Different sports have domain-specific semantic concepts, structures, and features. The challenge of autoencoding and transforming the domain-specific features and models to bridge the semantic gap for sports videos has yet to be overcome. The problem of knowledge interpolation between domains, called domain adaptation (DA), is an emerging research area in the field of machine learning. In brief, training and testing data may come from different domains. Determining how to solve the target-transfer learning problem remains challenging. Numerous approaches to DA have been presented [239], [240], [241]. The research paradigm in sports video analysis will face the same problem in the future. Learning with exchangeable information from different types of sports videos is very challenging and represents a promising research direction. For example, does a model trained using the previous year's data remain valid for adapting to a new season? Furthermore, despite the sports belonging to different categories, do the tactics in a soccer game remain valid for a baseball game? In addition, given the similarity in game rule, can the offense or defense in a table tennis game be applied to a tennis game?

### C. Robust Visual Feature Extracting for Precise Human Action Recognition

With the invention of high-precision sensors and vision-based action acquisition techniques, precise body features can be obtained for modeling key human objects by using an effective and high-accuracy scheme. Currently, action recognition has been a mainstream research topic not only in sports [94], [138], [242], but also in movies [243], healthcare [92], video surveillance [244], and building monitoring [245]. A direction for research extension is to develop a robust visual feature extraction scheme for understanding human actions in detail. A sports action video typically depicts multiple human actions simultaneously. To interpret a complex video, an appropriate approach is to adopt multiple dense detailed labels, instead of a single description [242]. By using robust feature extraction approaches, researchers can accurately analyze the context of a sport. Regarding video dynamics, Li *et al.* [250] encoded dynamics of deep features for action recognition. They found that explicit modeling of long-range dynamics is more important than short- and medium-range ones for action recognition.

Recently, several contributions have focused on extracting precise trajectories of objects, such as dense trajectories [63], improved trajectories [95], and multiobject trajectory using spatiotemporal convolution kernels [246] to determine human actions in detail. Ma *et al.* [251] used rich feature hierarchies of CNNs as target object representations and learned a linear correlation filter on each CNN layer to improve tracking accuracy and robustness. When these methods are used in tracking video objects, an intelligent system is required to perform deeper information discovery and to determine the relationships between trajectory data and actions. A recent review article [247] reported the impact of revamped deep neural networks on action recognition. For sports applications, deep learning was used to analyze basketball trajectories and predict whether a three-point shot is successful [248]. In addition, deep neural networks were used to construct classifiers that can recognize NBA offensive plays [249].

Although we have already seen numerous examples of successful applications of feature extraction and action recognition methods, many open problems remain because of the diversity of game structures among sports domains. Developing a unified framework that enables processing data from diverse sports is still challenging. The tradeoff between commonality and robustness must be to overcome. Ultimately, the prospective goal of action recognition in sports is to develop a machine that can read, write, listen to, and speak a voice-over to broadcast sports videos directly.

## V. CONCLUSIONS

We conducted a comprehensive survey of existing approaches on sports content analysis. Several approaches were reviewed from the viewpoint of content-aware scalability. First, we introduced the fundamentals of content analysis, such as the concept of the content pyramid, the categorization of the sports genre, and an overview of the sports video analytics. Second, we reviewed the state-of-the-art studies conducted in this decade according to the content hierarchical model. The methods of content-aware analysis were discussed with respect to object-, event-, and context-oriented groups. Finally, we reported the prominent challenges identified in the literature and categorized the potential future directions in sports video content analysis. We believe that our survey can advance the field of research on content-aware video analysis for sports.

## ACKNOWLEDGEMENTS

This work was partially supported by Ministry of Science and Technology of Taiwan, under grant MOST105-2221-E- 155-066, and Innovation Center for Big Data and Digital Convergence, under grant No. 228196. The author would like to thank the associate editor and reviewers for their valuable comments and the authors that contributed figures to this survey.

## REFRENCE

[1] M. Marchi and J. Albert. *Analyzing Baseball Data with R*. Chapman and Hall/CRC Press, Taylor & Francis Group (Oct. 29, 2013).
[2] J. Albert and J. Bennett. *Curve Ball: Baseball, Statistics, and the Role of Chance in the Game*. Copernicus; Softcover reprint of the original 1st ed. 2001 edition (April 8, 2003)
[3] T. Tango, M. Lichtman, and A. Dolphin. *The Book: Playing The Percentages In Baseball*. CreateSpace Independent Publishing Platform (April 28, 2014).
[4] S. M. Shea and C. E. Baker. *Basketball Analytics: Objective and Efficient Strategies for Understanding How Teams Win*. CreateSpace Independent Publishing Platform (November 5, 2013).
[5] S. M. Shea. *Basketball Analytics: Spatial Tracking*. CreateSpace Independent Publishing Platform (December 7, 2014)
[6] D. Oliver. Basketball on Paper: Rules and Tools for Performance Analysis. Potomac Books (November 1, 2004).
[7] SAP Web site. [Online]. Available: http://go.sap.com/index.html
[8] Vizrt Web site. [Online]. Available: http://www.vizrt.com
[9] Vizrt in Wikipedia Web site. [Online]. Available: http://en.wikipedia.org/wiki/Vizrt
[10] MLBAM Web site. [Online]. Available: http://www.mlbam.com
[11] Sportvision Web site. [Online]. Available: http://www.sportvision.com
[12] Brooks Baseball Web site. [Online]. Available: http://www.brooksbaseball.net
[13] Sportradar Web site. [Online]. Available: http://sportradar.com
[14] J. R. Wang and N. Parameswaran, "Survey of Sports Video Analysis: Research Issues and Applications," in *Proc. Pan–Sydney Area Workshop Vis. Info. Process.*, 2003, pp. 87–90.
[15] A. Kokaram, N. Rea, R. Dahyot, A. M. Tekalp, P. Bouthemy, P. Gros, and I Sezan, "Browsing Sports Videos: Trends in Sports-related Indexing and Retrieval Work," *IEEE Signal Process. Magazine*, vol. 23, no. 2, pp. 47–58, pp. Mar. 2006.
[16] C.B. Santiago, A. Sousa, M. L. Estriga, L. P. Reis, and M. Lames, "Survey on Team Tracking Techniques Applied to Sports," in *Proc. IEEE Int'l Conf. Autonomous Intell. Syst.*, 2010, pp. 1–6.
[17] T. D'Orazio and M. Leo, "A review of vision-based systems for soccer video analysis," *Pattern Recogn.*, vol. 43, no. 8, pp. 2911–2926, Aug. 2010.

[18] A. Rehman and T. Saba, "Features extraction for soccer video semantic analysis: current achievements and remaining issues," *Artif. Intell. Rev.*, vol. 41, no. 3, pp. 451–461, 2014.

[19] S.F. de Sousa Júnior, A. de A. Araújo, and D. Menotti "An overview of automatic event detection in soccer matches," in *Proc. IEEE Workshop Applicant. Comput. Vis.*, 2011, pp. 31–38.

[20] J. Han, D. Farin, and P. H. N. de With, "Broadcast Court-Net Sports Video Analysis Using Fast 3–D Camera Modeling," *IEEE Trans. Circuits Syst. Video Technol.*, vol. 18, no. 11, pp. 1628–1638, Nov. 2008.

[21] H. C. Shih and C. L. Huang, "Content-based Scalable Video Retrieval System," in *Proc. IEEE Int'l Symp. Circuits Syst.*, 2005, pp. 1553–1556.

[22] H. C. Shih and C. L. Huang, "Content-based multi–functional video retrieval System," in *Proc. IEEE Int'l Conf. Consum. Electron.*, 2005, pp. 383–384.

[23] D. Brezeale and D. J. Cook, "Automatic Video Classification: A Survey of the Literature," *IEEE Trans. Syst. Man Cybernetics—PART C: Applicant. Reviews*, vol. 38, no. 3, pp. 416–430, May 2008.

[24] J. Wang, C. Xu, and E. Ching, "Automatic sports video genre classification using pseudo–2d–HMM," in *Proc. IEEE Int'l Conf. Pattern Recogn.*, 2006, pp. 778–781.

[25] X. Gibert, H. Li, and D. Doermann, "Sports video classification using HMMs," in *Proc. IEEE Int'l Conf. Multimedia Expo.*, 2003, pp. 345–348.

[26] A. Ekin and A. M. Tekalp, "Shot Type Classification by Dominant Color for Sports Video Segmentation and Summarization," in *Proc. IEEE Int'l Conf. Acoustics, Speech and Signal Process.*, 2003, pp. 173–176.

[27] E. Jaser, J. Kittler, and W. Christmas, "Hierarchical decision making scheme for sports video categorization with temporal post-processing," in *Proc. IEEE Conf. Comput. Vis. Pattern Recogn.*, 2004, pp. 908–913.

[28] X. Yuan, W. Lai, T. Mei, X. S. Hua, X. Q. Wu, and S. Li, "Automatic video genre categorization using hierarchical SVM," in *Proc. IEEE Int'l Conf. Image Process.*, 2006, pp. 2905–2908.

[29] L. Li, N. Zhang, L. Y. Duan, Q. Huang, J. Du, and L. Guan, "Automatic Sports Genre Categorization and View-type Classification over Large-scale Dataset," in *Proc. ACM Multimedia*, 2009, pp. 653–656.

[30] J. You, G. Liu, and A. Perkis, "A Semantic framework for video genre classification and event analysis," *Signal Process.: Image Commun.*, vol. 25, no. 4, pp. 287–302, 2010.

[31] L. Y. Duan, M. Xu, Q. Tian, C. S. Xu, and J. S. Jin, "A Unified Framework for Semantic Shot Classification in Sports Video," *IEEE Trans. multimedia*, vol. 7, no. 6, pp. 1066–1083, Dec. 2005.

[32] N. Zhang, L. Y. Duan, L. Li, Q. Huang, J. Du, W. Gao, and L. Guan, "A Generic Approach for Systematic Analysis of Sports Videos," *ACM Trans. Intell. Syst. and Technol.*, vol. 3, no. 3, article 46, May 2012.

[33] F. Cricri, M. J. Roininen, J. Leppanen, S. Mate, I. D. D. Curcio, S. Uhlmann, and m. Gabbouj, "Sport Type Classification of Mobile Videos," *IEEE Trans. Multimedia*, vol. 16, no. 4, pp. 917–932, June 2014.

[34] M. Sugano, T. Yamada, S. Sakazawa, S. Hangai, "Genre Classification Method for Home Videos," in *Proc. IEEE Int'l Workshop Multimedia Signal Process.*, 2009, pp. 1–5.

[35] L. Breiman, "Random Forests," *Mach. Learn.*, vol. 45, no. 1, pp. 5–32, 2001.

[36] L. Breiman, "Bagging Predictors," *Mach. Learn.*, vol. 24, no. 2, pp. 123–140, 1996.

[37] M. Naphade, T. Kristjansson, B. Frey, and T. S. Huang, "Probabilistic multimedia object (multijects): A novel approach to indexing and retrieval in multimedia systems," in *Proc. Fifth IEEE Int'l Conf. Image Process.*, Oct 1998, pp. 536–540.

[38] Y. Luo, T. D. Wu, J. N. Hwang, "Object-based analysis and interpretation of human motion in sports video sequences by dynamic Bayesian networks," *Comput. Vis. Image Understand.*, vol. 92, no. 2–3, pp. 196–216, 2003.

[39] V. Pallavia, J. Mukherjeea, Arun K. Majumdara, and Shamik Suralb, "Ball detection from broadcast soccer videos using static and dynamic features," *J. Vis. Commun. Image R.*, vol. 19, no. 7, pp.426–436, Oct. 2008.

[40] M. Leo, P. L. Mazzeo, M. Nitti, and P. Spagnolo, "Accurate ball detection in soccer images using probabilistic analysis of salient regions," *Machine Vis. Applicant.*, vol. 24, no. 8, pp. 1561–1574, 2013.

[41] Y.C. Jiang, K.T. Lai, C. H. Hsieh, and M. F. Lai, "Player detection and tracking in broadcast tennis video," in *Proc. Third Pacific Rim Sym., PSIVT 2009*, LNCS 5414, pp. 759–770, 2009.

[42] H. T. Chen, C. L. Chou, W. J. Tsai, S. Y. Lee, and J. Y. Yu, "Extraction and representation of human body for pitching style recognition in broadcast baseball video," in *Proc. IEEE Int'l Conf. Multimedia Expo.*, July 2011, pp.1–4.

[43] K. L. Hua, C. T. Lai, C. W. You, and W. H. Cheng, "An efficient pitch-by-pitch extraction algorithm through multimodal information," *Info. Sci.*, vol. 294, pp. 64–77, 2015.

[44] R. Hamid, R. K. Kumar, M. Grundmann, K. Kim, I. Essa, and J. Hodgins, "Player Localization Using Multiple Static Cameras for Sports Visulaization," in *Proc. IEEE Comput. Vis. Pattern Recogn.*, 2010, pp. 731–738.

[45] Y. Liu, D. Liang, Q. Huang, and W. Gao, "Extracting 3D information from broadcast soccer video," *Image Vis. Comput.*, vol. 24, no. 10, pp.1146–1162, Oct. 2006.

[46] Y. Liu, D. Liang, Q. Huang, and W. Gao, "Self-calibration Based 3D Information Extraction and Application in Broadcast Soccer Video," in *Proc. Asia Conf. Comput. Vis.*, 2006, LNCS 3852, pp. 852–861.

[47] L. J. Li, H. Su, Y. Lim, and L. Fei-Fei, "Object Bank: An Object-Level Image Representation for High-Level Visual Recognition," *Int. J. Comput. Vis.*, vol. 107, no. 1, pp. 20–39, March 2014.

[48] H. C. Shih, J. N. Hwang, and C. L. Huang, "Content-based Video Attention Ranking using Visual and Contextual Attention Model for Baseball videos," *IEEE Trans. Multimedia*, vol. 11, no. 2, pp. 244–255, Feb. 2009.

[49] M. Wischnewski, A. Belardinelli, W. Schneider, and J. Steil, "Where to Look Next? Combining Static and Dynamic Proto-objects in a TVA-based Model of Visual Attention," *Cognit. Comput.*, vol. 2, no. 4, pp. 326–343, 2010.

[50] C. Bundesen, "A theory of visual attention," *Psychol. Rev.*, vol. 97, no. 4, pp. 523–547, 1990.

[51] S. H. Khatoonabadi and M. Rahmati, "Automatic soccer players tracking in goal scenes by camera motion elimination," *Image Vis. Comput.*, vol. 27, no. 4, pp.469–479, Mar. 2009.

[52] J. Liu, X. Tong, W. Li, T. Wang, Y. Zhang, and H. Wang, "Automatic player detection, labeling and tracking in broadcast soccer video," *Pattern Recogn. Lett.*, vol. 30, no. 2, pp.103–113, Jan. 2009.

[53] M. Kristan, J. Perš, M. Perše, and S. Kovačič, "Closed-world tracking of multiple interacting targets for indoor–sports applications," *Comput. Vis. Image Understand.*, vol. 113, no. 5, pp.598–611, May 2009.

[54] L. Sha, P. Lucey, S. Morgan, and D. Pease, "Swimmer Localization from a Moving Camera," in *Proc. IEEE Int'l Conf. Digital Image Comput.: Technol. Applicant.*, 2013, pp. 1–8.

[55] B. Chakraborty and S. Meher, "A Real–time Trajectory–based Ball Detection–and–tracking Framework for Basketball Video," *J. Opt.*, vol. 42, no. 2, pp. 156–170, April–June 2013.

[56] G. J. Liu, X.L. Tang, H. D. Cheng, J. H. Huang, and J. F. Liu, "A novel approach for tracking high speed skaters in sports using a panning camera," *Pattern Recogn.*, vol. 42, no. 11, pp. 2922–2935, Nov. 2009.

[57] J. Miura, T. Shimawaki, T. Sakiyama, and Y. Shirai, "Ball route estimation under heavy occlusion in broadcast soccer video," *Comput. Vis. Image Understand.*, vol. 113, no. 5, pp.653–652, May 2009.

[58] J. Liu, P. Carr, R. T. Collins, and Y. Liu, "Tracking Sports Players with Context-Conditioned Motion Models," in *Proc. IEEE Comput. Vis. Pattern Recogn.*, 2013, pp. 1830–1837.

[59] F. Yan, A. Kostin, W. Christmas, and J. Kittler, "A Novel Data Association Algorithm for Object Tracking in Clutter with Application to Tennis Video Analysis," in *Proc. IEEE Conf. Comput. Vis. Pattern Recogn.*, 2006, pp. 634–641.

[60] F. Yan, W. Christmas, and J. Kittler, "Layered Data Association Using Graph-Theoretic Formulation with Applications to Tennis Ball Tracking in Monocular Sequences," *IEEE Trans. Pattern Anal. Mach. Intell.*, vol. 30, no. 10, pp. 1814–1830, Oct. 2008.

[61] X. Zhou, L. Xie, Q. Huang, S. J. Cox, and Y. Zhang, "Tennis Ball Tracking Using a Two-Layered Data Association Approach," *IEEE Trans. Multimedia*, vol. 17, no. 2, pp. 145–156, Feb. 2015.

[62] H. Wang, A. Klaser, C. Schmid, and C.L. Liu, "Action Recognition by Dense Trajectories," in *Proc. IEEE Comput. Vis. Pattern Recogn.*, 2011, pp. 3169 – 3176.

[63] H. Wang, A. Kläser, C. Schmid, and C.L. Liu, "Dense Trajectories and Motion Boundary Descriptors for Action Recognition," *Int'l J. of Comput. Vis.*, vol. 103, no. 1, pp. 60–79, May 2013.

[64] Z. Niu, X. Gao, and Q. Tian, "Tactic analysis based on real-world ball trajectory in soccer video," *Pattern Recogn.*, vol. 45, no. 5, pp.1937–1947, May 2012.

[65] H. T. Chen, C. L. Chou, T. S. Fu, S. Y. Lee, and B. S. P. Lin, "Recognizing tactic patterns in broadcast basketball video using player trajectory," *J. Vis. Commun. Image R.*, vol. 23, no. 6, pp. 932–947, Aug. 2012.

[66] P. Lucey, A. Bialkowski, P. Carr, Y. Yue, and I. Matthews, "How to Get an Open Shot: Analyzing Team Movement in Basketball using Tracking Data," in *Proc. MIT SSAC*, 2014.

[67] X. Yu, C. Xu, H. W. Leong, Q. Tian, Q. Tang, and K. W. Wan, "Trajectory-Based Ball Detection and Tracking with Applications to Semantic Analysis of Broadcast Soccer Video," in *Proc. ACM Multimedia*, 2003, pp.11–20.

[68] A. Hervieu, P. Bouthemy, and J. P. Le Cadre, "A Statistical Video Content Recognition Method Using Invariant Features on Object Trajectories," *IEEE Trans. Circuits Syst. Video Technol.*, vol. 18, no. 11, pp. 1533–1543, Nov. 2008.

[69] A. Bialkowski, P. Lucey, P. Carr, Y. Yue, and I. Matthews, "Win at home and draw away: Automatic formation analysis highlighting the differences in home and away team behaviors," in *Proc. MIT SLOAN Sports Analytics Conf.*, 2014.

[70] Y. Zheng, "Trajectory Data Mining: An Overview," *ACM Trans. Intell. Syst. Technol.*, vol. 6, no. 3, article 29, May 2015.

[71] J. Vales–Alonso, D. Chaves–Dieguez, P. Lopez–Matencio, J. J. Alcaraz, F. J. Parrado–Garcia, and F. J. Gonzalez–Castano, "SAETA: A Smart Coaching Assistant for Professional Volleyball Training," *IEEE Trans. Syst. Man Cybernetics: Syst.*, vol. 45, no. 8, pp. 1138–1150, Aug. 2015.

[72] K. Choi and Y. Seo, "Automatic initialization for 3D soccer player tracking," *Pattern Recogn. Lett.*, vol. 32, no. 9, pp. 1274–1282, July 2011.

[73] X. Yu, N. Jiang, L. F. Cheong, H. W. Leong, and X. Yan, "Automatic camera calibration of broadcast tennis video with applications to 3D virtual content insertion and ball detection and tracking," *Comput. Vis. Image Understand.*, vol. 113, no. 5, pp. 643–652, May 2009.

[74] P. J. Figueroa, N. J. Leite, and R. M. L. Barros, "Tracking soccer players aiming their kinematical motion analysis," *Comput. Vis. Image Understand.*, vol. 101, no. 2, pp.122–135, Feb. 2006.

[75] J. Ren, J. Orwell, G. A. Jones, and M. Xu, "Tracking the soccer ball using multiple fixed cameras," *Comput. Vis. Image Understand.*, vol. 113, no. 5, pp.633–642, May 2009.

[76] S. Satoh, Y. Nakamura, and T. Kanade, "Name-It: Naming and Detecting Faces in News Videos," *IEEE Multimedia*, vol. 6, no. 1, pp. 22–35, 1999.

[77] M. Everingham, J. Sivic, and A. Zisserman, "Taking the bite out of automated naming of characters in TV video," *Image Vis. Comput.*, vol. 27, no. 5, pp. 545–559, 2009.

[78] V. Ramanathan, A. Joulin, P. Liang, and L. Fei-Fei, "Linking People in Videos with "Their" Names Using Coreference Resolution," in *Proc. Eur. Conf. Comput. Vis.*, LNCS 8689, 2014, pp. 95–110.

[79] W. L. Lu, J. A. Ting, K. P. Murphy, and J. J. Little, "Identifying Players in Broadcast Sports Videos using Conditional Random Fields," in *Proc. IEEE Comput. Vis. Pattern Recogn.*, 2011.

[80] A. Bialkowski, P. Lucey, P. Carr, Y. Yisong, S. Sridharan, and I. Matthews, "Large-scale analysis of Soccer matches using spatiotemporal tracking data," in *Proc. IEEE Conf. Data Mining*, vol., no., pp.725,730, 14–17 Dec. 2014.

[81] N. Nitta, N. Babaguchi, and T. Kitahashi, "Extracting actors, actions and events from sports video –a fundamental approach to story tracking," in *Proc. IEEE Int'l Conf. Pattern Recogn.*, 2000, pp. 718–721.

[82] T. Sato, T. Kanade, E. K. Hughes, M. A. Smith, and S. Satoh, "Video OCR: indexing digital news libraries by recognition of superimposed captions," *Multimedia Syst.*, vol. 7, no. 5, pp. 385–395, 1999.

[83] S. Messelodi and C. M. Modena, "Automatic identification and skew estimation of text lines in real scene images," *Pattern Recogn.*, vol. 32, no. 5, pp. 791–810, 1999.

[84] J. J. Weinman, E. Learned–Miller, and A. R. Hanson, "Scene text recognition using similarity and a lexicon with sparse belief propagation," *IEEE Trans. Pattern Anal. Mach Intell.*, vol. 31, no. 10, pp. 1733–1746, 2009.

[85] T. Yamamoto, H. Kataoka, M. Hayashi, Y. Aoki, K. Oshima, and M. Tanabiki, "Multiple players tracking and identification using group detection and player number recognition in sports video," in *Proc. IEEE IECON*, 2013, pp. 2442–2446.

[86] A. Pnevmatikakis, N. Katsarakis, P. Chippendale, C. Andreatta, S. Messelodi, C. M. Modena, and F. Tobia, "Tracking for context extraction in athletic events," in *Proc. Int'l workshop social, adaptive and personalized multimedia interaction and access, ACM Multimedia*, 2010, pp 67–72.

[87] C. Patrikakis, A. Pnevmatikakis, P. Chippendale, M. Nunes, R. Santos Cruz, S. Poslad, W. Zhenchen, N. Papaoulakis, and P. Papageorgiou, "Direct your personal coverage of large athletic events," *IEEE MultiMedia*, vol. 18, no. 4, pp. 18–29, 2011.

[88] S. Messelodi and C. M. Modena, "Scene text recognition and tracking to identify athletes in sport videos," *Multimedia Tools Applicant.*, vol. 63, no. 2, pp. 521–545, Mar. 2013.

[89] S. Gerke and K. Müller, "Soccer Jersey Number Recognition Using Convolutional Neural Networks," in *Proc. IEEE Int'l Conf. Comput. Vis.*, 2015, pp. 17–24.

[90] M. B. Holte, C. Tran, M. M. Trivedi, and T. B. Moeslund, "Human Pose Estimation and Activity Recognition from Multi-View Videos: Comparative Explorations of Recent Developments," *IEEE J. Selected Topics Signal Process.*, vol. 6, no. 5, pp. 538–552, Sept. 2012.

[91] D. Weinland, R. Ronfard, and E. Boyer, "A Survey of Vision–based Methods for action representation, segmentation and recognition," *INRIA Rep.*, vol. RR–7212, pp. 54–111, 2010.

[92] S. R. Ke, H. L. U. Thuc, Y. J. Lee, J. N. Hwang, J. H. Yoo, and K. H. Choi, "A review on Video–based Human Activity Recognition," *Computers*, vol. 2, no. 2, pp. 88–131, 2013.

[93] G. Guo and A. Lai, "A survey on still image based human action recognition," *Pattern Recogn.*, vol. 47, no. 10, pp. 3343–3361, Oct. 2014.

[94] L. Wang, Y. Qiao, and X. Tang, "Action Recognition with Trajectory-Pooled Deep-Convolutional Descriptors," in *Proc. IEEE Comput. Vis. Pattern Recogn.*, 2015.

[95] H. Wang and C. Schmid, "Action recognition with improved trajectories," in *Proc. IEEE Int'l Conf. Comput. Vis.*, 2013.

[96] K. Simonyan and A. Zisserman, "Two-stream convolutional networks for action recognition in videos," in *Proc. NIPS*, 2014.

[97] B. Yao and L. Fei-Fei, "Modeling mutual context of object and human pose in human–object interaction activities," in *Proc. IEEE Comput. Vis. Pattern Recogn.*, 2010, pp. 17–24.

[98] B. Yao, L. Fei-Fei, "Recognizing human-object interactions in still images by modeling the mutual context of objects and human poses," *IEEE Trans. Pattern Anal. Mach. Intell.*, vol. 32, no. 9, pp. 1691–1703, Sept. 2012.

[99] P. Pecev, M. Racković, and M. Ivković, "A System for Deductive Prediction and Analysis of Movement of Basketball Referees," *Multimedia Tools Applicant.*, vol. 75, no. 23, pp. 16389–16416, 2016.

[100] H. Ghasemzadeh and R. Jafari, "Coordination Analysis of Human Movements With Body Sensor Networks: A Signal Processing Model to Evaluate Baseball Swings," *IEEE Sensors J.*, vol. 11, no. 3, pp. 603–610, March 2011.

[101] A. Schmidt, "Movement Pattern Recognition in Basketball Free-throw Shooting," *Human Movement Science*, vol. 31, no. 2, pp. 360–382, April 2012.

[102] H. Miyamori and S. Iisaku, "Video Annotation for Content-based Retrieval using Human Behavior Analysis and Domain Knowledge," in *Proc. IEEE Int'l Conf. Autom. Face Gesture Recogn.*, 2000, pp. 320–325.

[103] V. Kazemi, M. Burenius, H. Azizpour, and J. Sullivan, "Multi-view Body Part Recognition with Random Forest," in *Proc. BMVC 2013*, Sept. 2013.

[104] M. Burenius, J. Sullivan and S. Carlsson, "3D Pictorial Structures for Multiple View Articulated Pose Estimation," in *Proc. IEEE Comput. Vis. Pattern Recogn.*, June 2013, pp. 3618 – 3625.

[105] E. Swears, A. Hoogs, J. Qiang, and K. Boyer, "Complex Activity Recognition Using Granger Constrained DBN (GCDBN) in Sports and Surveillance Video," in *Proc. IEEE Comput. Vis. Pattern Recogn.*, 2014, PP.788–795.

[106] D. Tao, X. Li, X. Wu, and S. Maybank, "General tensor discriminant analysis and gabor features for gait recognition," *IEEE Trans. Pattern Anal. Mach. Intell.*, vol. 29, no. 10, pp. 1700–1715, 2007.

[107] N. Dalal and B. Triggs, "Histograms of oriented gradients for human detection," in *Proc. IEEE Int'l Conf. Pattern Recogn.*, 2005, pp. 886–893

[108] K. Mikolajczyk and C. Schmid, "A performance evaluation of local descriptors," *IEEE Trans. Pattern Anal. Mach. Intell.*, vol. 27, no. 10, pp. 1615–1630, 2005.

[109] M.A.R. Ahad, J.K. Tan, H. Kim, and S. Ishikawa, "Motion history image: its variants and applications," *Machine Vis. Applicant.*, vol. 23, pp. 255–281, 2012.

[110] D. Weinland, R. Ronfard, and E. Boyer, "Free viewpoint action recognition using motion history volumes," *Comput. Vis. Image Understand.*, vol. 104, no. 2, pp. 249–257, 2006.

[111] A. Klaser, M. Marszalek, and C. Schmid, "A spatio-temporal descriptor based on 3d gradients," in *Proc. British Machine Vis. Association*, 2008, pp. 995–1004.

[112] I. Mikic, M. M. Trivedi, E. Hunter, and P. Cosman, "Human body model acquisition and tracking using voxel data," *Int. J. Comput. Vis.*, vol. 53, no. 3, pp. 199–223, 2003.

[113] M. Hofmann and D. Gavrila, "Multi-view 3D human pose estimation in complex environment," *Int'l J. Comput. Vis.*, vol. 96, no. 1, pp. 103–124, 2011.

[114] Q. Delamarre and O. Faugeras, "3D articulated models and multiview tracking with physical forces," *Comput. Vis. Image Understand.*, vol. 81, no. 3, pp. 328–357, 2001.

[115] G. Cheung and T. Kanade, "A real-time system for robust 3d voxel reconstruction of human motions," in *Proc. IEEE Comput. Vis. Pattern Recogn.*, 2000.

[116] S. Y. Cheng and M. M. Trivedi, "Articulated human body pose inference from voxel data using a kinematically constrained Gaussian mixture model," in *Proc. IEEE Comput. Vis. Pattern Recogn. Workshop*, 2007.

[117] B. C. Shen, H. C. Shih, and C. L. Huang, "Real-time human motion capturing system," in *Proc. IEEE Int'l Conf. Image Process.*, 2005, pp. 1322–1325.

[118] S. R. Ke, J. N. Hwang, K. M. Lan, and S. Z. Wang, "View-Invariant 3D Human Body Pose Reconstruction using a Monocular Video Camera," in *Proc. Fifth ACM/IEEE Int'l Conf. Distributed Smart Cameras*, 2011, pp. 1–6.

[119] S. Zhang, H. Yao, X. Sun, K. Wang, J. Zhang, X. Lu, and Y. Zhang, "Action Recognition based on overcomplete independent components analysis," *Info. Sci.*, vol. 281, pp. 635–647, Oct. 2014.

[120] Q. Le, W. Zou, S. Yeung, A. Ng, "Learning hierarchical invariant spatio-temporal features for action recognition with independent subspace analysis," in *Proc. IEEE Int'l Conf. Pattern Recogn.*, 2011, pp. 3361–3368.

[121] W. Zhou, C. Wang, B. Xiao, and Z. Zhang, "Action Recognition via Structured Codebook Construction," *Signal Process.: Image Commun.*, vol. 29, no. 4, pp. 546–555, April 2014.

[122] H. Li, J. Tang, S. Wu, and Y. Zhang, "Automatic Detection and Analysis of Player Action in Moving Background Sports Video Sequences," *IEEE Trans. Circuits Syst. Video Technol.*, vol. 20, no. 3, pp. 351–364, Mar. 2010.

[123] G. Taylor, R. Fergus, Y. Lecun, C. Bregler, "Convolutional learning of spatio-temporal features," in *Proc. 11th Eur. Conf. Comput. Vis.*, 2010, pp. 140–153.

[124] R. Memisevic and G. Hinton, "Unsupervised learning of image transformations," in *Proc. IEEE Comput. Vis. Pattern Recogn.*, 2007.

[125] J. M. Chaquet, E. J. Carmona, and A. Fernandez–Caballero, "A survey of video datasets for human action and activity recognition," *Comput. Vis. Image Understand.*, vol. 117, no. 6, pp. 633–659, June 2013.

[126] M. D. Rodriguez, J. Ahmed, and M. Shah, "Action MACH a spatio-temporal maximum average correlation height filter for action recognition," in *Proc. IEEE Comput. Vis. Pattern Recogn.*, 2008, pp.1–8.

[127] UCF Sports Web site. [Online]. Available: http://crcv.ucf.edu/data/UCF_Sports_Action.php

[128] H. Wang, M. Ullah, A. Klaser, I. Laptev, and C. Schmid, "Evaluation of local spatio-temporal features for action recognition," in *Proc. British Machine Vis. Conf.*, 2009, pp. 124.1–124.11.

[129] A. Kovashka and K. Grauman, "Learning a hierarchy of discriminative space-time neighborhood features for human action recognition," in *Proc. IEEE Comput. Vis. Pattern Recogn.*, 2010, pp. 2046–2053.

[130] Q.V. Le, W.Y. Zou, S.Y. Yeung, and A.Y. Ng, "Learning hierarchical invariant spatio-temporal features for action recognition with independent subspace analysis," in *Proc. IEEE Comput. Vis. Pattern Recogn.*, 2011, pp. 3361–3368.

[131] S. Sadanand, and J. Corso, "Action bank: a high-level representation of activity in video," in *Proc. IEEE Comput. Vis. Pattern Recogn.*, 2012, pp. 1234–1241.

[132] T. de Campos, M. Barnard, K. Mikolajczyk, J. Kittler, F. Yan, W. Christmas, and D. Windridge, "An evaluation of bags–of–words and spatio–temporal shapes for action recognition," in *Proc. IEEE Workshop Applicant. Comput. Vis.*, Jan. 2011, pp. 344–351.

[133] A. Khan, D. Windridge, and J. Kittler, "Multi-Level Chinese Takeaway Process and Label–Based Processes for Rule Induction in the Context of Automated Sports Video Annotation," *IEEE Trans. Cybernetics*, vol. 44, no. 10, pp. 1910–1923, Oct. 2014.

[134] ACASVA Dataset. [Online]. Available: http://www.cvssp.org/acasva/Downloads

[135] KTH Computer Vision Group Web site. [Online]. Available: http://www.csc.kth.se/cvap/cvg/

[136] Y. Wang, H. Jiang, M. S. Drew, Z.N. Li, and G. Mori, "Unsupervised discovery of action classes," in *Proc. IEEE Comput. Vis. Pattern Recogn.*, June 2006, pp. 1654–1661.

[137] SFU VM Lab. [Online]. Available: http://www2.cs.sfu.ca/research/groups/VML/people_cluster/

[138] A. Karpathy, G. Toderici, S. Shetty, T. Leung, R. Sukthankar, and L. Fei-Fei, "Large-scale Video Classification with Convolutional Neural Networks," in *Proc. IEEE Comput. Vis. Pattern Recogn.*, 2014.

[139] Sorts-1M-dataset [Online]. Available: https://github.com/gtoderici/sports-1m-dataset

[140] J. Y. H. Ng, M. Hausknecht, S. Vijayanarasimhan, O. Vinyals, R. Monga, and G. Toderici, "Beyond Short Snippets: Deep Networks for Video Classification," in *Proc. IEEE Comput. Vis. Pattern Recogn.*, 2015.

[141] J. C. Niebles, C. W. Chen, and Li Fei-Fei, "Modeling Temporal Structure of Decomposable Motion Segments for Activity Classification," in *Proc. 11th Eur. Conf. Comput. Vis.*, 2010.

[142] Stanford Olympic Sports Dataset. [Online]. Available: http://vision.stanford.edu/Datasets/OlympicSports/

[143] S. M. Safdarnejad, X. Liu, L. Udpa, B. Andrus, J. Wood, and D. Craven, "Sports Videos in the Wild (SVW): A Video Dataset for Sports Analysis," in *Proc. IEEE Int'l Conf. and Workshops Autom. Face and Gesture Recogn.*, 2015, pp. 1–7.

[144] MSU SVW Dataset. [Online]. Available: http://www.cse.msu.edu/~liuxm/sportsVideo

[145] Y. Gong, L. T. Sin, C. H. Chuan, H. Zhang, and M. Sakauchi, "Automatic Parsing of TV Soccer Programs," in *Proc. IEEE Int'l Conf. Multimedia Comput. Syst.*, 1995, pp.167–174.

[146] D. A. Sadlier, N. O'Connor, S. Marlow, and N. Murphy, "A Combined Audio-visual Contribution to Event Detection in Field Sports Broadcast Video. Case Study: Gaelic Football," in *Proc. 3rd IEEE Int'l Symp. Signal Process. Info. Technol.*, 2003, pp.552–555.

[147] C.H. Liang, W.T. Chu, J.H. Kuo, J.L. Wu, and W.H. Cheng, "Baseball Event Detection Using Game-Specific Feature Sets and Rules," in *Proc. IEEE Int'l Symp. Circuits Syst.*, 2005, pp.3829–3832.

[148] P. Chang, M. Han, and Y. Gong, "Extract Highlights from Baseball Game Video with Hidden Markov Models," in *Proc. IEEE Int'l Conf. Image Process.*, 2002, pp. 609–612.

[149] C. Poppe, S. De Bruyne, and R. V. de Walle, "Generic Architecture for Event Detection in Broadcast Sports Video," in *Proc. 3rd Int'l workshop Automated info. extraction media production*, 2010, pp.51–56.

[150] D. A. Sadlier, and N. E. O'Connor, "Event Detection in Field Sports Video Using Audio-Visual Features and a Support Vector Machine," *IEEE Trans. Circuits Syst. Video Technol.*, vol. 15, no. 10, pp.1225–1233, Oct. 2005.

[151] C.Y. Chen, J.C. Wang, J.F. Wang, and Y.H. Hu, "Event-Based Segmentation of Sports Video Using Motion Entropy," in *Proc. IEEE Int'l Symp. Multimedia*, 2007, pp.107–111.

[152] C. Wu, Y.F. Ma, H.J. Zhang, and Y.Z. Zhong, "Events Recognition by Semantic Inference for Sports Video," in *Proc. IEEE Int'l Conf. Multimedia Expo.*, 2002, pp.805–808.

[153] B. Li and M. I. Sezan, "Event Detection and Summarization in Sports Video," in *Proc. IEEE workshop CBAIVL*, 2001, pp.132–138.

[154] L. Xie, P. Xu, S.F. Chang, A. Divakaran, and H. Sun, "Structure Analysis of Soccer Video with Domain Knowledge and Hidden Markov Models," *Pattern Recogn. Lett.*, vol. 25, no. 7, pp.767–775, May 2004.

[155] D. W. Tjondronegoro, and Y.P. Phoebe Chen, "Knowledge-Discounted Event Detection in Sports Video," *IEEE Trans. Syst., Man Cybernetics, Part A: Syst. Humans*, vol. 40, no. 5, pp.1009–1024, Sept. 2010.

[156] C. Liu, Q. Huang, S. Jiang, L. Xing, Q. Ye, and W. Gao, "A Framework For Flexible Summarization of Racquet Sports Video Using Multiple Modalities," *Comput. Vis. Image Understand.*, vol.113, no. 3, pp.415–424, March 2009.

[157] G. Miao, G. Zhu, S. Jiang, Q. Huang, C. Xu, and W. Gao, "A Real-Time Score Detection and Recognition Approach For Broadcast Basketball Video," in *Proc. IEEE Int'l Conf. Multimedia Expo.*, 2007, pp.1691–1694.

[158] S. Nepal, U. Srinivasan, and G. Reynolds, "Automatic Detection of 'Goal' Segments in Basketball Videos," in *Proc. ACM Multimedia*, 2001, pp.261–269.

[159] M.H. Hung, and C.H. Hsieh, "Event Detection of Broadcast Baseball Videos," *IEEE Trans. Circuits Syst. Video Technol.*, vol. 18, no. 12, pp.1713–1726, Dec. 2008.

[160] Y. Rui, A. Gupta, and A. Acero, "Automatically Extracting Highlights for TV Baseball Programs," in *Proc. ACM Multimedia*, 2000, pp.105–115.

[161] Z. Xiong, R. Radhakrishnan, A. Divakaran, and T. S. Huang, "Audio Events Detection Based Highlights Extraction from Baseball, Golf and Soccer Games in A Unified Framework," in *Proc. IEEE Int'l Conf. Multimedia Expo.*, 2003, pp.401–404.

[162] H.C. Shih, and C.L. Huang, "Detection of The Highlights in Baseball Video Program," in *Proc. IEEE Int'l Conf. Multimedia Expo.*, 2004, pp.595–598.

[163] M. H. Kolekar, "Bayesian belief network based broadcast sports video indexing," *Multimedia Tools Applicant.*, vol. 54, no. 1, pp. 27–54, Aug. 2011.

[164] M. H. Kolekar and S. Sengupta, "Bayesian Network-Based Customized Highlight Generation for Broadcast Soccer Videos," *IEEE Trans. Broadcast.*, vol. 61, No. 2, pp.195–209, June 2015.

[165] M. Petkovic, V. Mihajlovic, W. Jonker, and S. Djordjevic–Kajan, "Multi-model Extraction of Highlights from TV Formula 1 Programs," in *Proc. IEEE Int'l Conf. Multimedia Expo.*, 2002, pp.817–820.

[166] C.Y. Chao, H.C. Shih, and C.L. Huang, "Semantic-based Highlight Extraction of Soccer Program Using DBN," in *Proc. IEEE Int'l Conf. Acoustics, Speech Signal Process.*, 2005, pp. 1057–1060.

[167] Y. Gong, M. Han, W. Hua, and W. Xu, "Maximum Entropy Model-based Baseball Highlight Detection and Classification," *Comput. Vis. Image Understand.*, vol. 96, no. 2, pp.181–199, Nov. 2004.

[168] X. Qian, H. Wang, G. Liu, and X. Hou, "HMM based Soccer Video Event Detection Using Enhanced Mid–level Semantic," *Multimedia Tools Applicant.*, vol. 60, no. 1, pp.233–255,

[169] N. Babaguchi, Y. Kawai, T. Ogura, and T. Kitahashi, "Personalized Abstraction of Broadcasted American Football Video by Highlight Selection," *IEEE Trans. Multimedia*, vol. 6, no. 4, pp.575–586, Aug. 2004.

[170] J. Assfalg, M. Bertini, C. Colombo, A. Del Bimbo, and W. Nunziati, "Semantic Annotation of Soccer Videos: Automatic Highlights Identification," *Comput. Vis. Image Understand.*, vol. 92, no. 2–3, pp.285–305, Nov. 2003.

[171] M. Lazarescu, and S. Venkatesh, "Using Camera Motion to Identify Types of American Football Plays," in *Proc. IEEE Int'l Conf. Multimedia Expo.*, 2003, pp. 181–184.

[172] R. A. Sharma, V. Gandhi, V. Chari, and C. V. Jawahar, "Automatic analysis of broadcast football videos using contextual priors," *SIViP* 2016.

[173] G. Kobayashi, H. Hatakeyama, K. Ota, Y. Nakada, T. Kaburagi, and T. Matsumoto, "Predicting viewer-perceived activity/dominance in soccer games with stick-breaking HMM using data from a fixed set of cameras," *Multimedia Tools Applicant.*, vol. 75, no. 6, pp. 3081–3119, Mar. 2016.

[174] H.T. Chen, H.S. Chen, and S.Y. Lee, "Physics-based Ball Tracking in Volleyball Videos with Its Applications to Set Type Recognition and Action Detection," in *Proc. IEEE Int'l Conf. Acoustics, Speech Signal Process.*, 2007, pp. 1097–1100.

[175] F. Cricri, S. Mate, I.D.D. Curcio, and M. Gabbouj, "Salient Event Detection in Basketball Mobile Videos," in *Proc. IEEE Int'l Symp. Multimedia*, 2014, pp. 63–70.

[176] W.L. Lu, K. Okuma, and J. J. Little, "Tracking and Recognizing Actions of Multiple Hockey Players Using the Boosted Particle Filter," *Image Vis. Comput.*, vol. 27, no. 1–2, pp.189–205, Jan. 2009.

[177] H. Pan, P. van Beek, and M. I. Sezan, "Detection of Slow-Motion Replay Segments in Sports Video for Highlights Generation," in *Proc. IEEE Int'l Conf. Acoustics, Speech Signal Process.*, 2001, pp.1649–1652.

[178] M.L. Shyu, Z. Xie, M. Chen, and S. C. Chen, "Video Semantic Event/Concept Detection Using a Subspace-Based Multimedia Data Mining Framework," *IEEE Trans. Multimedia*, vol. 10, no. 2, pp. 252–259, Feb. 2008.

[179] G. Zhu, Q. Huang, C. Xu, L. Xing, W. Gao, and H. Yao, "Human Behavior Analysis for Highlight Ranking in Broadcast Racket Sports Video," *IEEE Trans. Multimedia*, vol. 9, no. 6, pp.1167–1182, Oct. 2007.

[180] C. L. Huang, H. C. Shih, and C. Y. Chao, "Semantic Analysis of Sports Video using Dynamic Bayesian Network," *IEEE Trans. Multimedia*, vol. 8, no. 4, pp749–760, Aug. 2006.

[181] M. Tavassolipour, M. Karimian, and S. Kasaei, "Event Detection and Summarization in Soccer Videos Using Bayesian Network and Copula," *IEEE Trans. Circuits Syst. Video Technol.*, vol. 24, no. 2, pp. 291–304, Feb. 2014.

[182] R. B. Nelsen. An Introduction to Copulas. 2nd ed. New York: Springer, 2006, pp. 7–14.

[183] C. K. Chow and C. N. Liu, "Approximating discrete probability distributions with dependence trees," *IEEE Trans. Info. Theory*, vol. 14, no. 3, pp. 462–467, May 1968.

[184] D. Zhong, and S.F. Chang, "Real-time View Recognition and Event Detection for Sports Video," *J. Vis. Commun. Image R.*, vol. 5, no. 3, pp.330–347, Sep. 2004.

[185] K. Choro, "Automatic Detection of Headlines in Temporally Aggregated TV Sports news Videos," *IEEE Int'l Symp. Image Signal Process. Anal.*, 2013, pp. 147–153.

[186] H. T. Chen, C. L. Chou, W. C. Tsai, S. Y. Lee, and B. S. P. Lin, "HMM-based ball hitting event exploration system for broadcast baseball video," *J. Vis. Commun. Image R.*, vol. 23, no. 5, pp. 767–781, July 2012.

[187] W. Zhou, A. Vellaikal, and C. C. J. Kuo, "Rule-based Video Classification System for Basketball Video Indexing," in *Proc. ACM Multimedia*, 2000, pp.213–216.

[188] M. Chen, S.C. Chen, M.L. Shyu, and K. Wickramaratna, "Semantic Event Detection via Multimodal Data Mining," *IEEE Signal Process. Magazine*, vol. 23, no. 2, pp.38–46, March 2006.

[189] G. Zhu, C. Xu, Q. Huang, Y. Rui, S. Jiang, W. Gao, and H. Yao, "Event Tactic Analysis Based on Broadcast Sports Video," *IEEE Trans. Multimedia*, vol. 11, no.1, pp.49–67, Jan. 2009.

[190] T. D'Orazio, M. Leo, P. Spagnolo, M. Nitti, N. Mosca, and A. Distante, "A Visual System for Real Time Detection of Goal Events During Soccer Matches," *Comput. Vis. Image Understand.*, vol. 113, no. 5, May 2009.

[191] T. Saba, and A. Altameem, "Analysis of Vision Based Systems to Detect Real Time Goal Events in Soccer Videos," *Int'l J. Appl. Artif. Intell.*, vol. 27, no. 7, pp.656–667, Aug 2013

[192] C. Snoek, and M. Worring, "Time Interval Maximum Entropy Based Event Indexing in Soccer Video," in *Proc. IEEE Int'l Conf. Multimedia Expo.*, 2003, pp. 481–484.

[193] J. Nichols, J. Mahmud, and C. Drews, "Summarizing Sporting Events Using Twitter," in *Proc. ACM Intell. User Interfaces*, 2012, pp.189–198.

[194] N. Babaguchi, Y. Kawai and T. Kitahashi, "Event Based Indexing of Broadcasted Sports Video by Intermodal Collaboration," *IEEE Trans. Multimedia*, vol. 4, no. 1, pp.68–75, Mar 2002.

[195] A. Gupta, P. Srinivasan, J. Shi, and L. S. Davis, "Understanding videos, constructing plots learning a visually grounded storyline model from annotated videos," in *Proc. IEEE Comput. Vis. Pattern Recogn.*, 2009, pp. 2012–2019.

[196] Y.M. Su and C.H. Hsieh, "A Novel Model-based Segmentation Approach to Extract Caption Contents on Sports Videos," in *Proc. IEEE Int'l Conf. Multimedia Expo.*, 2006, pp.1829–1832.

[197] J. Guo, C. Gurrin, S. Lao, C. Foley, and A. F. Smeaton, "Localization and Recognition of the Scoreboard in Sports Video Based on SIFT Point Matching," in *Proc. ACM MMM*, 2011, LNCS 6524, 2011, pp. 337–347.

[198] X. Tang, X. Gao, J. Liu, and H. Zhang, "A Spatial-Temporal Approach for Video Caption Detection and Recognition," *IEEE Tran. Neural Networks*, vol.13, no.4, pp.961–971, Nov. 2002.

[199] D. Zhang, R. K. Rajendran, and S. F. Chang, "General and Domain Specific Techniques for Detecting and Recognizing Superimposed Text in Video," in *Proc. IEEE Int'l Conf. Image Process.*, 2002, pp.593–596.

[200] S. H. Sung and W. S. Chun, "Knowledge-Based Numeric Open Caption Recognition for Live Sportscast," in *Proc. IEEE Int'l Conf. Pattern Recogn.*, 2002, pp.822–825

[201] R. Lienhart. Video OCR: A survey and practitioner's guide. In *Video Mining*, Kluwer Academic Publisher, pp. 155–184, 2003.

[202] S. Mori, C. Y. Suen, and K. Yamamoto, "Historical review of OCR research and development," *Proceed. IEEE*, vol. 80, pp. 1029–1058, July 1992.

[203] G. J. Vanderbrug and A. Rosenfeld, "Two-Stage Template Matching," *IEEE Trans. Computers*, vol. 26, no. 4, pp. 384–393, April 1997.

[204] H. C. Shih and K. C. Yu, "SPiraL Aggregation Map (SPLAM): A new descriptor for robust template matching with fast algorithm," *Pattern Recogn.*, vol. 48, no. 5, pp. 1707–1723, May, 2015.

[205] W. N. Lie and S. H. Shia, "Combining Caption and Visual Features for Semantic Event Classification of Baseball Video," in *Proc. IEEE Int'l Conf. Multimedia Expo.*, 2005, pp.1254–1257.

[206] H. C. Shih and C. L. Huang, "Content Extraction and Interpretation of Superimposed Captions for Broadcasted Sports Videos," *IEEE Trans. Broadcast.*, vol.54, no.3, pp.333–346, Sept. 2008.

[207] H. C. Shih and C. L. Huang, "Semantics Interpretation of Superimposed Captions in Sports Videos," in *Proc. IEEE Workshop Multimedia Signal Process.*, Oct. 2007, pp.235–238.

[208] C. Jung and J. Kim, "Player Information Extraction for Semantic Annotation in Golf Videos," *IEEE Trans. Broadcast.*, vol. 55, no. 1, pp. 79–83, March 2009.

[209] N. Steinmetz, "Context-aware Semantic Analysis of Video Metadata," Ph. D dissertation, University of Potsdam, May 2014.
[210] S. P. Yong, J. D. Deng, and M. K. Purvis, "Wildlife video key-frame extraction based on novelty detection in semantic context," *Multimedia Tools Applicant.*, vol. 62, no. 2, pp. 359–376, 2013.
[211] H. C. Shih, "A Novel Attention-Based Key-Frame Determination Method," *IEEE Trans. Broadcast.*, vol. 59, no. 3, pp. 556–562, Sept. 2013.
[212] L. Wu, Y. Gong, X. Yuan, X. Zhang, and L. Cao, "Semantic aware sport image resizing jointly using seam carving and warping," *Multimedia Tools Applicant.*, vol. 70, no. 2, pp. 721–739, 2014.
[213] M. F. Weng and Y. Y. Chuang, "Cross–Domain Multicue Fusion for Concept–Based Video Indexing," *IEEE Trans. Pattern Anal. Mach. Intell.*, vol. 34, no. 10, pp. 1927–1941, Oct. 2012.
[214] H. C. Shih, J. N. Hwang, and C. L. Huang, "Content-based Video Attention Ranking using Visual and Contextual Attention Model for Baseball videos," *IEEE Trans. Multimedia*, vol. 11, no. 2, pp. 244–255, Feb. 2009.
[215] H. C. Shih, C. L. Huang, and J. N. Hwang, "An Interactive Attention-Ranking System for Video Search," *IEEE MultiMedia*, vol. 16, no. 4, pp. 70–80, Oct.–Dec. 2009.
[216] C.Y. Chiu, P.C. Lin, S.Y. Lin, T.H. Tsai, and Y.L. Tsai, "Tagging Webcast Text in Baseball Videos by Video Segmentation and Text Alignment," *IEEE Trans. Circuits Syst. Video Technol.*, vol. 22, no. 7, pp. 999–1013, 2012.
[217] M. Jaderberg, K. Simonyan, A. Vedaldi, and A. Zisserman, "Reading Text in the Wild with Convolutional Neural Networks," *Int'l J. Comput. Vis.*, vol. 116, pp. 1–20, 2016.
[218] G. Li, S. Ma, and Y. Han, "Summarization-based Video Caption via Deep Neural Networks," in *Proc. ACM Multimedia*, 2015, pp. 1191–1194.
[219] S. Venugopalan, H. Xu, J. Donahu, M. Rohrbach, R. Mooney, and K. Saenko, "Translating Videos to Natural Language Using Deep Recurrent Neural Networks," in *Proc. NAACL HLT*, 2015, pp. 1494–1504.
[220] X. Chen and A.L. Yuille, "Detecting and Reading Text in Natural Scenes," in *Proc. IEEE Comput. Vis. Pattern Recogn.*, 2004, pp.366–373.
[221] M. R. Lyu, J. Song, and M. Cai, "A Comprehensive Method for Multilingual Video Text Detection, Localization, and Extraction," *IEEE Trans. Circuits Syst. Video Technol.*, vol.15, no.2, pp.243–255, Feb. 2005.
[222] J. Xi, X.S. Hua, X.R. Chen, and L. Wenyin, "A Video Text Detection and Recognition System," in *Proc. IEEE Int'l Conf. Multimedia Expo.*, Aug. 2001, pp.873–876.
[223] R. Lienhart and A. Wernicke, "Localizing and Segmenting Text in Images and Videos," *IEEE Trans. Circuits Syst. Video Technol.*, vol.12, no.4, pp.256–268, Aug. 2002.
[224] D. Noll, M. Schwarzinger, and W. Seelen, "Contextual Feature Similarities for Model-Based Object Recognition," in *Proc. IEEE Int'l Conf. Comput. Vis.*, May 1993, pp.286–290.
[225] Z. Wang, J. Yu, and Y. He, "Soccer Video Event Annotation by Synchronization of Attack-Defense Clips and Match Reports with Coarse-grained Time Information," *IEEE Trans. Circuits Syst. Video Technol.*, in press, 2016.
[226] M.A. Refaey, W. Abd–Almageed, and L. S. Davis, "A Logic Framework for Sports Video Summarization using Text–Based Semantic Annotation," in *Proc. Int'l workshop Semantic Media Adaptation Personalization*, 2008, pp.69–75.
[227] Y. Zhang, C. Xu, Y. Rui, J. Wang, and Y. Zhang, "Semantic Event Extraction from Basketball Games Using Multi-Modal Analysis," in *Proc. IEEE Int'l Conf. Multimedia Expo.*, 2007, pp.2190–2193.
[228] D. Zhang and S. F. Chang, "Event detection in baseball video using superimposed caption recognition," in *Proc. ACM Multimedia*, 2002, pp. 315–318.
[229] C.M. Chen and L.H. Chen, "A novel approach for semantic event extraction from sports webcast text," *Multimedia Tools Applicant.*, vol.71, no.3, pp.1937–1952, Aug. 2014.
[230] C. Xu, Y. F. Zhang, G. Zhu, and Y. Rui, "Using Webcast Text for Semantic Event Detection in Broadcast Sports Video," *IEEE Trans. Multimedia*, vol.10, no.7, pp.1342–1355, Nov. 2008.
[231] N. Nitta, Y. Takahashi, and N. Babaguchi, "Automatic personalized video abstraction for sports videos using metadata," *Multimedia Tools Applicant.*, vol.41, no.1, pp.1–25, Jan. 2009.
[232] Cisco Visual Networking Index (VNI), "The Zettabyte Era: Trends and Analysis," White paper, June 10, 2014.
[233] Y. Jin, Y. Wen, and C. Westphal, "Optimal Transcoding and Caching for Adaptive Streaming in Media Cloud: an Analytical Approach," *IEEE Trans. Circuits Syst. Video Technol.*, vol. 25, no. 12, pp. 1914–1925, Dec. 2015.
[234] Y. Wu, T. Mei, Y. Q. Xu, N. Yu, and S. Li, "MoVieUp: Automatic Mobile Video Mashup," *IEEE Trans. Circuits Syst. Video Technol.*, vol. 25, no. 12, pp. 1941-1954, Dec. 2015.
[235] Y. Wen, X. Zhu, J. J. P. C. Rodrigues, and C. W. Chen, "Cloud Mobile Media: Reflections and Outlook," *IEEE Trans. Multimedia*, vol. 16, no. 4, pp. 885–902, June 2014.
[236] P. Hitzler and K. Janowicz, "Linked Data, Big Data and the 4th Paradigm," *Semantic Web Journal*, vol. 4, no. 3, pp. 233–235, July 2013.
[237] K. Mahmood and H. Takahashi, "Cloud Based Sports Analytics Using Semantic Web Tools and Technologies," in *Proc. IEEE Global Conf. Consum. Electron.*, *2015*, pp. 431–432.
[238] Y. Guo, Y. Liu, A. Oerlemans, S. Lao, S. Wu, and M. S. Lew, "Deep Learning for Visual Understanding: A review," *Neurocomputing*, vol. 187, pp. 27–48, Apr. 2016
[239] S. Chopra, S. Balakrishnan, and R. Gopalan, "DLID: Deep learning for domain adaptation by interpolating between domains," in *Proc. Int'l Conf. Mach. Learn.*, 2013.
[240] R. Gopalan, R. Li, and R. Chellappa, "Unsupervised Adaptation Across Domain Shifts By Generating Intermediate Data Representations", *IEEE Trans. Pattern Anal. Mach. Intell.*, vol. 36, pp. 2288–2302, Nov 2014.
[241] J. Donahue, Y. Jia, O. Vinyals, J. Hoffman, N. Zhang, E. Tzeng, and T. Darrell, "DeCAF: A Deep Convolutional Activation Feature for Generic Visual Recognition," in *Proc. Int'l Conf. Mach. Learn.*, pp. 647–655, 2014.
[242] S. Yeung, O. Russakovsky, N. Jin, M. Andriluka, G. Mori, and L Fei-Fei, "Every Moment Counts: Dense Detailed Labeling of Actions in Complex Videos," *arXiv preprint arXiv*:1507.05738, 2015.
[243] H. Wang, D. Oneata, J. Verbeek, and C. Schmid, "A Robust and Efficient Video Representation for Action Recognition," *Int'l J. Comput. Vis.*, vol. 119, pp. 219–238, 2016.
[244] C. B. Jin, S. Li, T. D. Do, and H. Kim, "Real-Time Human Action Recognition Using CNN Over Temporal Images for Static Video Surveillance Cameras," in *Proc. PCM 2015, vol. 9315 of the series Lecture Notes in Computer Science*, pp. 330–339.
[245] H. C. Shih, "Automatic Building Monitoring and Commissioning Via Human Behavior Recognition," in *Proc. IEEE Global Conf. Consum. Electron.*, 2016.
[246] K. Knauf, D. Memmert, and U. Brefeld, "Spatio-temporal Convolution Kernels," *Mach. Learn.*, vol. 102, pp. 247–273, 2016.
[247] S. Herath, M. Harandi, and F. Porikli, " Going Deeper into Action Recognition: A Survey," *arXiv preprint arXiv:1605.04988*, 2016.
[248] R. C. Shah and R. Romijnders, "Applying Deep Learning to Basketball Trajectories," *arXiv preprint arXiv:1608.03793*, 2016.
[249] K. C. Wang and R. Zemel, "Classifying NBA Offensive Plays Using Neural Networks," in *Proc. MIT SLOAN Sports Analytics Conf.*, 2016.
[250] Y. Li, W. Li, V. Mahadevan, and N. Vasconcelos, "VLAD$^3$: Encoding Dynamics of Deep Features for Action Recognition," in *Proc. IEEE Comput. Vis. Pattern Recogn.*, 2016, pp. 1951–1960.
[251] C. Ma, J. B. Huang, X. Yang, and M. H. Yang, "Hierarchical Convolutional Features for Visual Tracking," in *Proc. IEEE Int'l Conf. Comput. Vis.*, 2015, pp. 3074–3082.

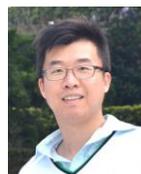

**Huang-Chia Shih (M'08)** received the B.S. degree with the highest honors in electronic engineering from the National Taipei University of Technology, Taipei, Taiwan in 2000 and the M.S. and Ph.D. degrees in electrical engineering (EE) from the National Tsing Hua University in 2002 and 2008, respectively.

He has been an associate professor at the department of EE, Yuan Ze University (YZU), Taoyuan, Taiwan, since 2016. His research interests are content-based multimedia processing, pattern recognition, and human-computer interaction (HCI). Dr. Shih served as visiting scholar in Department of EE, University of Washington from September 2006 to April 2007, visiting professor in John von Neumann Faculty of Informatics, Obuda University in Hungary in summer of 2011. He received Outstanding Youth Electrical Engineer Award from Chinese Institute of Electrical Engineering in December 2015, YZU Young Scholar Research Award from YZU in March 2015, Kwoh-Ting Li Young Researcher Award from ACM Taipei/Taiwan Chapter in April 2014, the Pan Wen Yuan Exploration Research Award from Pan Wen Foundation in May 2013, the best paper award of ISCE2013 from IEEE Consumer Electronics Society, and student paper award of GCCE2015 from CE Society. Dr. Shih is author and co-author of more than 50 papers in international journals and conferences. Dr. Shih is a member of IEEE.